\documentclass[journal,twocolumn]{IEEEtran}

\usepackage{float}
\usepackage{times}
\usepackage{subfigure}
\usepackage{epsfig}
\usepackage{graphicx}
\usepackage{amsmath}
\usepackage{amssymb}
\usepackage{bm}
\usepackage{color}

\usepackage[utf8]{inputenc}

\usepackage[pagebackref=true,breaklinks=true,colorlinks,bookmarks=false]{hyperref}

\graphicspath{{fig/}}

\newcommand{\bC}{{\mathbf C}}
\newcommand{\bH}{{\mathbf H}}

\newcommand{\bX}{{\mathbf X}}

\newcommand{\bU}{{\mathbf U}}
\newcommand{\bR}{{\mathbf R}}
\newcommand{\bg}{{\mathbf g}}
\newcommand{\bx}{{\mathbf x}}

\newcommand{\bbE}{{\mathbb E}}

\newcommand{\Ph}{P_{\mathrm{h}}}

\newcommand{\fixed}{I}
\newcommand{\moving}{V}

\DeclareMathOperator{\cov}{cov} \DeclareMathOperator{\var}{var}
\DeclareMathOperator{\tr}{tr}

\newcommand{\ave}{\mathrm{ave}}
\newcommand{\opt}{\mathrm{opt}}
\newcommand{\sel}{\mathrm{sel}}

\newtheorem{proposition}{Proposition}

\begin{document}

\title{Uncertainty driven probabilistic voxel selection for image registration}

\author{Boris~N.~Oreshkin,
        Tal~Arbel~\IEEEmembership{Member,~IEEE,}
\thanks{Copyright (c) 2013 IEEE. Personal use of this material is permitted. However, permission to use this material for any other purposes must be obtained from the IEEE by sending a request to pubs-permissions@ieee.org.}% <-this % stops a space
\thanks{Boris~N.~Oreshkin and Tal~Arbel are with the Centre for Intelligent Machines, Department
of Electrical and Computer Engineering, McGill University, QC, H3A 0E9, Canada (e-mail: boris.oreshkin@mail.mcgill.ca, arbel@cim.mcgill.ca).}% <-this % stops a space
\thanks{Manuscript received October 19, 2012; revised May 17, 2013.}}

%\author{name\thanks{blah}}

%\thanks{Copyright (c) 2013 IEEE. Personal use of this material is permitted. However, permission to use this material for any other purposes must be obtained from the IEEE by sending a request to pubs-permissions@ieee.org.}

%\author{Boris N. Oreshkin, Tal Arbel\\
%Centre for Intelligent Machines, McGill University\\
%Montreal, Quebec, Canada}

\maketitle

\begin{abstract}
  This paper presents  a  novel probabilistic voxel selection strategy {\color{black} for medical image registration in time-sensitive contexts}, where the goal is aggressive voxel sampling (e.g. using less than 1\% of the total number) while maintaining registration accuracy and low failure rate. We develop a Bayesian framework whereby, first, a voxel sampling probability field (VSPF) is built based on the uncertainty on the transformation parameters. We then describe a practical, multi-scale registration algorithm, where, at each optimization iteration, different voxel subsets are sampled based on the VSPF. The approach maximizes accuracy without committing to a particular fixed subset of voxels. The probabilistic sampling scheme developed is shown to manage the tradeoff between the robustness of traditional random voxel selection (by permitting more exploration) and the accuracy of fixed voxel selection (by permitting a greater proportion of informative voxels).
\end{abstract}

%%%%%%%%% BODY TEXT
\section{Introduction}
{\color{black}
Image registration is one of the critical problems in the field of medical imaging, spanning a wide range of applications including, but not limited to, diagnostic imaging~\cite{Wein08}, image guided interventions~\cite{Wein08,Zhang07,Sakai06}, image guided therapy~\cite{Webster09} and patient to atlas mapping~\cite{Chakravarty09}. } Typically, the evaluation of the similarity measure and/or its derivatives are required to perform the optimization over the transformation parameters. Many registration strategies have been developed under the implicit assumption that all of the voxels (or, at least, as many as necessary to achieve desirable result) can be used for calculating the similarity measure --- the ultimate goal being the highest accuracy at any computational cost. {\color{black} However, in time sensitive applications, such as in image guided interventions~\cite{Wein08,Zhang07,Sakai06} and image guided therapy~\cite{Webster09}, image registration must be performed both accurately and quickly in order for the techniques to be adopted by clinicians. An example of this type of domain is image guided neurosurgery, where ultrasound images acquired during the operation must be quickly matched to pre-operative MRI or CT in order to correct for brain {\color{black}shift~\cite{Wein08,Zhang07,Ji08,Roche01,DeNigris12}.} } Performing computations based on all the available image voxels can be prohibitively costly mainly due to the large number of voxel intensity values involved in the calculations. Time-sensitive applications generally benefit from techniques that speed up direct image registration by utilizing only a subset of available voxels during registration. In these contexts, a decrease in accuracy of several percentages can be tolerated in order to assure the preservation of robustness and a significant decrease in registration time. However, \emph{significant} speedups attained via a very aggressive reduction in the number of selected voxels (e.g. less than 1\% of the total number of voxels) often result in a significant deterioration in robustness (i.e. an increase in failure rate) and relatively rapid increase in the resulting registration error. {\color{black} In this paper, our main goal is to demonstrate the promise of a voxel sampling technique based on the Bayesian model for registration parameter uncertainty, by showing that a well designed sampling strategy is capable of maintaining the functionality of the registration procedure even if the average number of voxels used in similarity metric calculations is very low. In other words, we explore the limits to which image registration can be pushed in terms of voxel selection. Specifically, the goal is to develop a voxel selection (sampling) scheme that would work well even when a very small percentage of voxels (on the order of 0.1\%) is used for registration, thus attaining fast registration in contexts where time is the most valuable resource.  However, the registration framework based on this voxel selection scheme is also required to guarantee reliable and accurate results, a goal that is key to its adoption in the context of clinical practice (time sensitive domains, e.g. image guided therapy~\cite{Webster09} or image guided interventions~\cite{Wein08,Zhang07,Sakai06}). }

Randomized \emph{voxel selection} is one commonly used technique for reducing the computational cost of registration, where a randomly selected subset of all voxels are used to drive the optimization~\cite{Mattes2003,Viola1997}. This technique gained popularity due to its simplicity. { \color{black} Randomized voxel selection, which avoids any bias towards particular image patterns, does lead to a reduction in registration accuracy. Thus the utility of optimizing voxel selection has been well recognized. } Early works that attempted to optimize voxel selection presented heuristics for the relatively simple context of template matching~\cite{Barnea1972,Nagel1972} and demonstrated the advantage of optimized selection. More recently, a threshold-based optimized voxel selection method was proposed by Reevs and Hezar~\cite{Reeves1995}. They proposed a threshold based k-space voxel selection scheme based on the mean squared reconstruction error criterion in the context of the magnetic resonance image reconstruction. However, voxels selected by the threshold based approach tend to form spatially localized clusters leading to reduced robustness. Dallaert and Collins~\cite{Dellaert1999} attempted to overcome this effect and proposed a heuristic voxel selection strategy for the context of template based tracking. It involves computing the Jacobian of the similarity metric for every voxel and randomly selecting $M$ out of the top 20 percent voxels with the highest Jacobian~\cite{Dellaert1999}. Brooks and Arbel~\cite{Brooks2007} extended the approach of Dellaert and Collins~\cite{Dellaert1999} by proposing an information theoretic selection criterion and addressing the issue of the Jacobian scale. In practice, many methods simply favour voxels with high intensity gradient magnitudes since they better reveal internal structure of the image and thus are deemed more useful for obtaining a good alignment. Furthermore, Sabuncu and Ramadge proposed the gradient magnitude subsampling (GMS) approach, where the moving image is probabilistically subsampled using non-uniform grid generated based on the probabilities proportional to the gradient magnitude~\cite{Sabuncu2004}. This approach permits diversification and spread of the subsampled voxels while still focusing on image details. The same paper discusses the fact that the randomized subsampling is superior to the subsampling on a regular grid (decimation). An extension of this work employing the mixture of the previous approach with the uniformly random subsampling was proposed by Oreshkin and Arbel in~\cite{Oreshkin2012}. Here it is proposed to sample from the convex combination of the probability fields generated via the GMS and the uniformly random sampling approaches and the mixing parameter is identified from learning procedure using a small training dataset. Finally, curvlet based sampling, recently proposed by Freiman et al.~\cite{Freiman2010} was tested on Vanderbilt RIRE (Retrospective Image Registration Evaluation) dataset~\cite{Fitzpatrick1998} and resulted in approximately the same level of accuracy as the GMS approach~\cite{Sabuncu2004}. All these approaches work well in some contexts, but most fail in the low sampling rate scenario that is the focus of the current paper. Moreover, existing approaches often lack a direct link to the task at hand, namely, the estimation of the registration parameters. Overall, developing a formal, general stochastic voxel selection framework for low sampling rate image registration is still an open research topic.

In this paper, we propose a new probabilistic voxel selection strategy based on a multi-resolution voxel sampling probability field (VSPF), formed by the set of probabilities associated with image voxels. At every scale, we optimize the VSPF to minimize the transformation parameter uncertainty calculated from the Bayesian generative model for the approximate moving image. In the proposed framework, the most informative voxels are selected more frequently, while the exploration of the image continues via sampling different voxel subsets at each optimization iteration. { \color{black} The fact that each scale has its own VSPF allows to further exploit this feature by emphasizing different voxel subsets at different resolution scales.} We derive the analytical solution of the VSPF optimization problem and illustrate the adaptive nature and generality of the approach. We show that deterministic selection rules similar to e.g. one presented in Reevs and Hezar~\cite{Reeves1995} and the random voxel selection based on uniform probabilities of Wells et al.~\cite{Viola1997} can be cast as particular cases of our framework. Specifically, the probabilistic sampling scheme developed manages the tradeoff between the robustness of random voxel selection (by permitting more exploration) and the accuracy of fixed voxel selection (by permitting a greater proportion of informative voxels). Finally, we propose a practical algorithm based on our  framework and demonstrate significant registration performance gains through multi-modal (MRI-CT) image registration experiments with the popular, publicly available RIRE Vanderbilt dataset~\cite{Fitzpatrick1998}. One of the limitations of this dataset is that validation is restricted to rigid multi-modal registration only.
The rest of the paper is organized as follows. In Section~\ref{sec:Direct_Image_registration} we describe the relationship between direct image registration and voxel selection and introduce the notion of the VSPF. {\color{black} In Section~\ref{sec:Uncertainty} we define voxel utility, which is used to optimize the VSPF and to build a practical multi-scale algorithm in Sections~\ref{sec:Optimization} and~\ref{sec:proposed_algorihm}. Results of numerical experiments are then presented in Section~\ref{sec:Numerical_Experiments}. Section~\ref{sec:dicussion_of_findings} discusses our findings, limitations inherent to the proposed methodology and to the validation method.} Section~\ref{sec:Conclusions} concludes the paper.

%-------------------------------------------------------------------------

\section{Problem Statement}
\label{sec:Direct_Image_registration}

Direct parametric image registration is formulated for a reference image, $\fixed(\bx) = [\fixed_{1}(\bx_1), \ldots, \fixed_{N}(\bx_N)]$, and a transformed moving image, $\moving(\mathrm{T}_{\theta}(\bx)) =  [\moving_{1}(\mathrm{T}_{\theta}(\bx_1)), \ldots, \moving_{N}(\mathrm{T}_{\theta}(\bx_N))]$ both defined as collections of $N$ voxels.  For clarity and simplicity we denote the images as $\fixed = [\fixed_{1}, \ldots, \fixed_{N}]$ and $\moving =  [\moving_{1}, \ldots, \moving_{N}]$. Each of the voxel intensities is defined as a mapping from the coordinate space $\mathcal{X} \subseteq \mathbb{R}^d$ to the intensity space $\mathcal{I} \subseteq \mathbb{R}$, $\fixed_i, \moving_i : \mathcal{X} \rightarrow \mathcal{I}, i = 1 \ldots N$, where $d$ is the dimensionality of coordinate space. The transformation function, $\mathrm{T}_{\theta} : \mathcal{X} \rightarrow \mathcal{X}$, is characterized by the parameters $\theta \in \Theta$. Typical examples of transformation functions include affine transformation containing rotation, translation, scale and skew as special cases; and non-rigid transformation based on e.g. B-spline modeling that allows capturing non-linear free-form image transforms. The goal is to find the transformation parameters that maximize the similarity metric $D_N : \mathcal{I}^{N\times2} \rightarrow \mathbb{R}$ that establishes correspondence among voxels of the moving and the reference images and then maps the intensity values of the reference and the moving images into a single number representing the degree of image similarity. Widely used similarity metrics are mutual information~\cite{Viola1997} and normalized mutual information (NMI)~\cite{Studholme99}~\footnote{\color{black}These metrics have been popular in general literature as well as in the context of the RIRE dataset}.

The registration parameter optimization problem associated with similarity metric $D_N$ can be formally stated as:
\begin{align} \label{eqn:reg_problem_N}
\theta_{\opt} = \arg\max_{\theta \in \Theta} D_N[\fixed(\bx),
\moving(\mathrm{T}_{\theta}(\bx))].
\end{align}
One of the major challenges encountered when solving this problem is the fact that the computational cost of computing $D_N$ increases with $N$ (sometimes linearly and sometimes quadratically). This could be a major obstacle when using registration algorithms in computer assisted interventions and other real time or soft real time applications. One of the most popular approaches to alleviating this problem is voxel selection, when only a subset of voxels are used to calculate the approximation to the similarity metric. Formally, voxel selection can be defined with a selection operator $\mathcal{S}^M : \mathcal{I}^{N} \rightarrow \mathcal{I}^{M}$ reducing the dimensionality of image intensity space from $N$ to $M$,
\begin{align} \label{eqn:reg_problem_M}
\theta_{\sel} = \arg\max_{\theta \in \Theta}
D_M[\fixed(\bx), \mathcal{S}^M \circ \moving(\mathrm{T}_{\theta}(\bx))].
\end{align}
The approximate solution $\theta_{\sel}$ based on the correspondences of $M < N$ voxels is thus less computationally expensive.

One popular approach to solving~\eqref{eqn:reg_problem_M}, provided we have a sufficiently good initial guess $\theta_0$, is the iterative Gauss-Newton algorithm:
 \begin{align}
 \theta_{n+1} = \theta_{n} + \bH_M^{-1} \nabla_{\theta_n} D_M[\fixed(\bx), \mathcal{S}^M \circ\moving(\mathrm{T}_{\theta_n}(\bx))],
 \label{eqn:GaussNewtonIteration}
 \end{align}
where the Hessian $\bH_M$ of the similarity metric with respect to the transformation parameters is given by:
\begin{align}
\bH_M = \nabla_{\theta_n} \nabla_{\theta_n} D_M[\fixed(\bx), \mathcal{S}^M \circ \moving(\mathrm{T}_{\theta_n}(\bx))].
\end{align}

We propose to extend the voxel selection framework by introducing the optimized voxel sampling probability field (VSPF) $\mathcal{P} \subseteq [0, 1]^N$, $\mathcal{P} = [p_1, p_2, \ldots, p_N]$, where $p_i$ in the column vector $\mathcal{P}$ determines the probability that we select $i$-th voxel for the calculation of the similarity metric and $\sum_{i=1}^N p_i$ is the average number of selected voxels. Here each $p_i$ is related to the randomized voxel selection indicator $d_i \in \{0, 1\}$ such that a given voxel is selected if $d_i=1$, discarded if $d_i=0$ and the expectation of the indicator is equal to $p_i$, $\bbE\{d_i\} = p_i$. We also propose to use a unique voxel selection $\mathcal{S}^M_n$ at every optimization iteration $n$ obtained by sampling from the VSPF. This results in the continued exploration of the images being registered. In the following sections we investigate how the notion of VSPF is related to the parameter uncertainty and develop a framework for optimizing the VSPF.

{\color{black}

\section{VSPF Optimization Based on Bayes Uncertainty} \label{sec:VSP_optimization_based_on_bayes_uncertainty}

In this section, we present a generative model and a set of assumptions underlying our analysis of the utility of each voxel for the registration procedure and then optimize the VSPF based on the utilities of voxels. One of the core foundations underlying our model is the Baysian approach. Within the Bayesian framework we assume that the transformation parameters are random variables. The natural justification for this assumption is the fact that the values of the registration parameters realized for a given pair of images depend on a large variety of factors that are impossible or hard to model explicitly. Uncertainty of patient positioning in both modalities and equipment calibration errors are just two of those factors. The most general and rigorous way to handle this uncertainty is to assume that the multiple unknown factors affecting the registration parameters and thus the registration parameters themselves are stochastic (random). The fact that there are many stochastic factors affecting the registration parameters suggests that one could choose a Gaussian distribution as an appropriate prior distribution for the registration parameters in the Bayesian model.

In Section~\ref{sec:Uncertainty} we create a generative model for the voxels of the moving image based on the Taylor expansion of the voxel values with respect to the registration transformation parameters. This model basically quantifies how a small perturbation in a registration parameter vector distorts the given voxel. Based on this model and on the Bayesian formulation of the registration parameter estimation problem, given voxels from the moving image, we quantify the utility of each voxel. In other words, we quantify the potential that each voxel has towards improving the estimation of registration parameters.

In Section~\ref{sec:Optimization} we optimize the VSPF in such a manner that the total registration parameter estimation error calculated from our generative model is minimized via proper assignment of VSPF values. We initially pose the VSPF optimization problem  as a constrained problem and then use Lagrange relaxation to find a simpler formulation. Our solution to the problem results in the VSPF that assigns higher sampling probabilities to the voxels that have higher utility. In other words, the proposed sampling strategy based on the optimized VSPF tends to favor those voxels that have better chance in helping to improve the estimation of registration parameters based on our generative model for the moving image.

Note that this section addresses exclusively the theoretical foundations of the proposed framework. Our implementation of these ideas as well as recommendations for setting the parameters are discussed in Section~\ref{sec:proposed_algorihm}.

}

\subsection{Analysis of Bayes Uncertainty}
\label{sec:Uncertainty}

In this section we describe the generative model that we use to assess the utility of each voxel during the voxel selection procedure. Generative models used to construct practical similarity metrics $D_N(\cdot)$ may be found e.g. in~\cite{Viola1997}. In this section we briefly state main analysis results without concentrating on technical details. {\color{black} The technical details (including the definitions of expectation and variance operators) are presented in Appendix~\ref{app:analysis_of_bayes_uncertainty}.}

We propose the following first-order Taylor approximation of the moving image $\moving_{i}(\mathrm{T}_{\theta}(\bx))$ defined in Section~\ref{sec:Direct_Image_registration}:
\begin{align} \label{eqn:generative_model}
\moving_{i}(\mathrm{T}_{\theta}(\bx)) \approx
\moving_{i}(\mathrm{T}_{\mu_{\theta}}(\bx)) +
\bg_{i}^T(\theta-\mu_{\theta}) + \xi_i,
\end{align}
where $\bg_{i} = \frac{\partial \moving_{i}(\mathrm{T}_{\theta}(\bx))}{\partial \mathrm{T}_{\theta}(\bx)} \frac{\partial \mathrm{T}_{\theta}(\bx)}{\partial\theta}\big|_{\theta = \mu_{\theta}}$ and $\xi_i$ is modeling noise.

{\color{black}
Our generative model and the ensuing analysis of Bayes uncertainty are based on the following set of assumptions:
\label{as_A}
\renewcommand{\theenumi}{$\mathcal{A}_\arabic{enumi}$}
\renewcommand{\labelenumi}{\theenumi}
\begin{enumerate}
    \item \label{as_A1} Registration parameters $\theta$ are random variables with prior Gaussian distribution having mean $\mu_\theta$ and covariance matrix $\bR_{\theta\theta}$.
    \item \label{as_A2} The modeling noise $\xi$ is Gaussian zero-mean with variance $\sigma_{\xi}^2$, independent of $\theta$ and such that for voxels $i \neq j$ $\xi_i$ is independent of $\xi_j$.
    \item \label{as_A3} The cross terms of the covariance matrix $\bR_{\moving\moving}$ are negligible: $|\bg_{i}^T\bR_{\theta\theta} \bg_{j}| \ll \sqrt{ (\bg_{i}^T\bR_{\theta\theta}\bg_{i}  + \sigma_{\xi}^2) ( \bg_{j}^T\bR_{\theta\theta} \bg_{j} + \sigma_{\xi}^2)}, \forall i\neq j$.
\end{enumerate}

\renewcommand{\theenumi}{\arabic{enumi}.}
\renewcommand{\labelenumi}{\theenumi}
}

Now we introduce the set  $\mathcal{D} = [d_1, d_2, \ldots, d_N ]$, composed of the randomized voxel selection decisions $d_i \in \{0, 1\}$. This is related to the vector of selected voxels $\moving_{\mathcal{D}} = [\ldots \moving_{i}, \ldots]^T$, where $\moving_{i}$ is included in $\moving_{\mathcal{D}}$ if $d_i = 1$. Recall that the decision vector is related to the VSPF through expectation: $\mathbb{E}\{d_i\} = p_i$. In other words, the $i$-th element of VSPF defines the frequency of making the decision $d_i=1$ corresponding to selecting the $i$-th voxel for similarity metric evaluation.

{\color{black} Let the optimal Bayesian estimator $\widehat\theta_{\mathcal{D}}$ obtained using a realization of the random vector $\moving_{\mathcal{D}}$ of voxel selection decisions.  We chose the variance of this optimal Bayesian estimator as a natural characterization of the Bayesian uncertainty. Under our modeling assumptions listed above it can be shown (please refer to Appendix~\ref{app:analysis_of_bayes_uncertainty} for details) that this variance has the following form: }
\begin{align} \label{eqn:theta_var}
\var(\theta - \widehat\theta_{\mathcal{D}}) = \bR_{\theta \theta} -
\sum_{i=1}^N p_i \bR_{\theta \moving_{i}} \bR_{\moving_{i} \moving_{i}}^{-1}
\bR_{\theta \moving_{i}}^T,
\end{align}
where $\bR_{\moving_{i} \moving_{i}}
=\var(\moving_{i})$ and $\bR_{\theta \moving_{i}} = \cov(\theta,  \moving_{i})$. Each term in the sum weighted by $p_i$ represents the average contribution of every voxel towards decreasing the variance of estimating the transform parameters on the basis of known moving image values
$\{ \moving_{i} \}_{i=1}^N$.

\subsection{Optimization of VSPF}
\label{sec:Optimization}

We now optimize the VSPF by posing the problem of optimizing registration parameter uncertainty characterized by the trace of the error covariance matrix. {\color{black} We chose to characterize registration parameter vector uncertainty by the trace of the parameter covariance matrix, which is a common choice in statistical parameter estimation literature: for a given estimator, the trace of its error covariance matrix is equal to the mean squared estimation error (see e.g.~\cite{Anderson05}, Example~3.6). Consequently, based on expression~\eqref{eqn:theta_var} we introduce $U_i = \tr(\bR_{\theta \moving_{i}} \bR_{\moving_{i} \moving_{i}}^{-1} \bR_{\theta \moving_{i}}^T)$ and call this voxel utility.} This value characterizes the potential that each voxel has towards decreasing the variance of estimating the transform parameters on the basis of known moving image values $\{ \moving_{i} \}_{i=1}^N$ under model~\eqref{eqn:generative_model}.

Our optimization problem is constrained by the voxel sampling probability $p_i$ being non-negative, upper bounded by $0< \Ph \leq 1$ and is subject to the constraint on the average cost $C_{\ave}$ of processing the selected voxels ($C$ is the single voxel processing cost). One could consider assigning different processing costs to different voxels by introducing $C_i$ instead of $C$. At the same time, in practice we have the case where $C_i = C$. Moreover, there is no loss of generality in supposing that $C_i = C$ as our current results can be easily extended in the case of different voxel processing costs. Because of this and because it makes our derivations more succinct and simpler to understand, we prefer to assume that $C_i = C$. Mathematically this can be formulated as follows:
\begin{align}
\mathcal{P}^* &= \min_{\mathcal{P}} \tr \var(\theta -
\widehat\theta_{\mathcal{D}}) \nonumber \\
\text{such that \quad} p_i &\geq
0, \quad p_i \leq \Ph, \quad \sum_{i=1}^N p_i C = C_{\ave}. \label{eqn:optProblem}
\end{align}
Using Lagrange multipliers and skipping the constant term $\tr \bR_{\theta \theta}$ we obtain the following Lagrange function:
\begin{align}
\mathcal{J}(\mathcal{P}) &=  - \sum_{i=1}^N p_i U_i - \sum_{i=1}^N
\gamma_i p_i \nonumber \\
&+ \sum_{i=1}^N \eta_i (p_i-\Ph) + \lambda (C_{\ave} -
\sum_{i=1}^N p_i C). \label{eqn:pspfCost}
\end{align}
{\color{black} It can be shown (see Appendix~\ref{app:OptimizationSolution}) that the problem \eqref{eqn:optProblem} can be reformulated as follows using the Lagrange relaxation method~\cite{Everett63} and introducing the auxiliary variable $A > 0$:
\begin{align}
\mathcal{P}^* &= \min_{A, \mathcal{P}} \mathcal{J}(\mathcal{P}) \nonumber \\
\text{such that \quad} p_i \geq
0, \quad p_i &\leq \Ph, \quad \sum_{i=1}^N p_i C = C_{\ave}, \nonumber \\
p_i &= A(U_i + \gamma_i - \eta_i + \lambda C).
\end{align}
}
The solution of the reformulated problem has the following form:
\begin{align} \label{eqn:pspfSolution}
p_i^* = \left\{  \begin{matrix} A(U_i + \lambda^* C) & \text{ if } & 0 \leq A(U_i + \lambda^* C) \leq \Ph  \\
\Ph & \text{ if } & A(U_i + \lambda^* C) > \Ph  \\
0 & & \text{otherwise}
\end{matrix} \right.
\end{align}
Here the auxiliary variable $A$ should be determined from the reformulated minimization problem and we will later show how this can be accomplished by analyzing the properties of the cost function. The lagrange multiplier $\lambda^*$ is selected to satisfy the average processing cost constraint, ${\mathcal{P}^*}^T \bC = C_{\ave}$ (here $(\cdot)^T$ is the vector transpose operator and $\bC = [C, \ldots, C]$ is the vector of costs).

Equation~\eqref{eqn:pspfSolution} provides a general intermediate solution. This intermediate result has interesting implications. From the set of inequalities $0 \leq A(U_i + \lambda^* C) \leq \Ph$ in~\eqref{eqn:pspfSolution} we can see that for an arbitrary value of $A$ the voxel selection distribution $\mathcal{P}^*$ is comprised of three types of voxels defined by two thresholds: $-\lambda^* C_i$ and $\Ph A^{-1}-\lambda^* C_i$. If the utility, $U_i$, of a given voxel is below the first threshold this voxel is never selected ($p_i^* = 0$). If the utility $U_i$ of the voxel is between the two thresholds, the voxel may or may not be selected with probability proportional to its utility, $p_i^* = A(U_i + \lambda^* C)$. Finally, for voxels deemed very informative, for which utility $U_i$ exceeds the second threshold we have $p_i^* = \Ph$ and thus voxels from this class are most frequently selected.

In the following we provide two theoretical results that characterize (i) the behaviour of ${\mathcal{J}}(\mathcal{P}^*)$ as a function of the auxiliary variable $A$ and (ii) the behaviour of the left hand side of the average processing cost constraint equation as a function of the Lagrange multiplier $\lambda^*$. These results will help us in specifying the optimal value of $A$ and advising an efficient numerical procedure to find $\lambda^*$.

The optimal value of $A$ can be found by minimizing ${\mathcal{J}}(\mathcal{P}^*)$. The following proposition, whose proof can be found in Appendix~\ref{app:ProofOfProposition1}, establishes that the cost function ${\mathcal{J}}$ evaluated at the optimized solution $\mathcal{P}^*$ is a non-increasing function of $A$.

\begin{proposition} \label{prop:1}
${\mathcal{J}}(\mathcal{P}^*)$ is a monotonically non-increasing function of $A, \forall A \geq 0$.
\end{proposition}

This implies that the formal optimal solution of the voxel selection problem is thus achieved when $A\rightarrow\infty$. It is straightforward to deduce that it has the following form (recall that $\lambda^*$ is a function of $A$):
\begin{align}
p_i^\star = \left\{  \begin{matrix} \lim\limits_{A\to\infty} A(U_i + \lambda^* C) & \text{ if } & - \lambda^* C \rightarrow U_i  \\
\Ph & \text{ if } & U_i > -\lambda^* C  \\
0 & & \text{otherwise}
\end{matrix} \right.
\label{eqn:pspfForm}
\end{align}

Before presenting another important result leading to a procedure to efficiently compute the value of $\lambda^*$ we define $\varphi(\lambda^*) \triangleq {\mathcal{P}^*}^T \bC$.
\begin{proposition} \label{prop:2}
Function $\varphi(\lambda^*)$ is a monotonically non-decreasing function of $\lambda^*, \forall \lambda^*$.
\end{proposition}
The proof of this result appears in Appendix~\ref{app:ProofOfProposition2}. Since the function defining constraint equation ${\mathcal{P}^*}^T \bC = C_\ave$ is monotonic, we can use efficient algorithms, for example, golden ratio search to find $\lambda^*$.

% \subsection{Analysis of Results}

 Solution~\eqref{eqn:pspfForm}, obtained in the case $A\to\infty$, assigns the highest probability, $\Ph$, to all the voxels for which voxel utility is greater than the threshold $-\lambda^* C$. Additionally, there is exactly one voxel, whose probability $\lim\limits_{A\to\infty} A(U_i + \lambda^* C)$ is selected to exactly satisfy the constraint ${\mathcal{P}^*}^T \bC = C_{\ave}$. Note that without the contribution of this voxel ${\mathcal{P}^*}^T \bC$ can only be incremented by $\Ph C$ with each new voxel whose utility exceeds the threshold $-\lambda^* C$.

 According to our experiments, the value of the cost function saturates as $A\rightarrow \infty$. Hence, large increments of $A$ tend to result in small increments of the cost function. Thus, the value of the cost function is almost the same at the optimal solution and at a suboptimal solution obtained with $A < \infty$. Typically, voxel selection strategies that define a fixed voxel subset tend to reduce the optimization capture range, robustness to local minima and exploratory power of the sampling scheme i.e. its ability to visit different image regions during optimization. This is true even if the voxel selections are optimized. One of the reasons for this is that it is difficult to correctly estimate the true value (utility) of a given voxel for a registration procedure. For any voxel selection criterion it is safe to say that noise and modeling errors result in highly unreliable estimates of the true value that a given voxel has for the registration problem. As a result, fixing voxel selection based even on the "most optimal" criterion may have catastrophic results for the registration problem if voxel value estimate is erroneous. On the other hand, the uniformly random sampling might decrease registration accuracy, but is more robust, since sampling induces image exploration and does not imply any bias towards particular voxel subsets. The general solution~\eqref{eqn:pspfSolution} with $A < \infty$  results in a more diverse VSPF since it assigns non-zero probabilities to a greater variety of voxels. At the same time, according to this solution, the additional voxels assigned with non-zero probabilities are defined based on their utilities and their probabilities are proportional to the respective utilities. This unique feature has potential to endow a sampling scheme with (i) vast exploratory power, (ii) ability to pick voxels most useful for achieving an accurate registration, (iii) ability to avoid catastrophic registration errors due to voxel utility modeling and estimation artifacts. These properties would help a sampling scheme achieve the robustness of randomized sampling approaches and improve accuracy by utilizing a greater proportion of meaningful voxels for similarity metric calculations.

\begin{figure*}[t]
\centering
\begin{tabular}{|l|}
\hline ~\\

\begin{minipage}[l]{0.9\textwidth}

\begin{enumerate}
\item Calculate $K$-scale image pyramid for the moving and the reference images;
\item Pick initialization $\theta_0$ and set parameters $C_{\ave}, P_{\mathrm{h}}$ as discussed in Section~\ref{sec:proposed_algorihm};
\item {\color{black} For every scale {$k=K,K-1,\ldots,1$} do}
    \begin{enumerate}
    \item Calculate utilities $\{ U_i \}_{i=1}^N$ for all voxels of the moving image;
    \item Estimate VSPF $\mathcal{P}_k$ as discussed in Sections~\ref{sec:Optimization} and~\ref{sec:proposed_algorihm};
    \item For iteration {$n=1,2,\ldots$, {\it L}} do
        \begin{itemize}
            \item Generate new voxel selection by sampling
            $\mathcal{P}_k$;
            \item Calculate Jacobian and Hessian based on the voxel selection;
            \item Update parameters using trust region Gauss-Newton iteration;
        \end{itemize}
    \item Pick $\theta_0$ for the next scale based on the last iteration (account for scale change);
    \end{enumerate}
\end{enumerate}

\end{minipage}

~\\~\\\hline
\end{tabular}

\caption{High level summary of the proposed algorithm}
\label{fig:alg_proposed}
\end{figure*}

We thus propose a sampling scheme based on~\eqref{eqn:pspfSolution}, where $A$ is taken as small as possible to diversify VSPF and as large as necessary in the view of being able to satisfy the average cost constraint. Here the latter requirement is due to the fact that for too small values of $A$ no $\lambda^*$ exists that could satisfy constraint equation ${\mathcal{P}^*}^T \bC = C_{\ave}$. Another parameter that regulates the exploratory power of the VSPF is the upper bound on the probability values, $\Ph$. Lower values of this parameter tend to favour image exploration. This parameter could be chosen to be a universal constant whose reasonable value could be fixed based on experience. We also propose a more accurate strategy to learn the value of this parameter based on a small training dataset. This way it could be adapted to the problem at hand, optimization strategy and other specifics of the registration system. In the following section, where we describe a practical multi-scale algorithm, we discuss both variants.

Finally, let us examine how our approach relates to other approaches discussed in the literature. Consider the case where $A \rightarrow \infty$, $\Ph = 1$, $C_{\ave}$ is an integer number and all the values of $U_i$ are unique (recall that $N$ is the total number of voxels in the image). According to~\eqref{eqn:pspfForm}, all the VSPF values will be either 0 or 1, i.e. voxels are either selected or ignored. In such context, the proposed voxel sampling methodology is similar in spirit to the selection strategy proposed by Reevs and Hezar~\cite{Reeves1995}, i.e. calculate some measure of voxel utility and compare it to a threshold. Consider an alternative case where $\Ph = 1$ and all $U_i$ values are equal (voxels are deemed equally important/informative). According to~\eqref{eqn:pspfForm}, the optimal solution then is to assign every voxel equal probability and $p_i^* = \lim\limits_{A\to\infty} A(U_i + \lambda^* C) = C_\ave / N$. In other words, such context yields an optimal voxel selection policy similar to the random uniform sampling proposed in~\cite{Viola1997}.

\section{Practical Design Guidelines}
\label{sec:proposed_algorihm}

We now describe the full practical implementation of our voxel sampling scheme, whose high-level algorithmic summary is outlined in Fig.~\ref{fig:alg_proposed}. The first stage involves creating an image pyramid of $K$-scales, where each level involves low-pass filtering and subsampling the original reference and moving images. Optimization is then performed in order of scale. Hence, the level with coarsest scale is optimized first, and the result of each optimization is fed as an initial point to the following level.

The VSPF is calculated from the moving image once at each scale using initial transformation parameter values and following the framework described in previous sections. In practice VSPF calculations can be performed either on the fixed or on the moving image, whichever is easiest. The choice mainly depends on the interpolation strategy (partial volume or linear/cubic) and the point at which voxels are sampled (in the moving image coordinate frame or in the fixed image coordinate frame). The computation of VSPF can be performed in different registration contexts. For example, in the context of the image guided neurosurgery one of the images (e.g. MRI) is often available before an operation and VSPFs can be computed ahead of time using parameter initializations corresponding to the identity transformation (no registration) to save time during registration. At the same time, if data are not available ahead of time, the proposed framework can be applied online, once per scale, (the way it is depicted in Fig.~\ref{fig:alg_proposed}), since the computational expense of computing the VSPF is small and consists of matrix manipulations with spatial image gradients and transform derivative matrices.

\subsection{Parameter Selections} \label{ssec:setting_Ph}

Voxel utility, $U_i$, is computed according to its definition in Section~\ref{sec:Uncertainty} as $U_i = (\bg_{i}^T \bR_{\theta\theta} \bg_{i} + \sigma^2_{\xi})^{-1} \tr(\bR_{\theta,\moving_i} \bR_{\theta,\moving_i}^T),$ where $\bR_{\theta,\moving_i} = \bR_{\theta\theta} \bg_i$. Recall that $\bg_{i} = \frac{\partial \moving_{i}(\mathrm{T}_{\theta}(\bx))}{\partial \mathrm{T}_{\theta}(\bx)} \frac{\partial \mathrm{T}_{\theta}(\bx)}{\partial\theta}\big|_{\theta = \mu_{\theta}}$ is the moving image intensity derivative with respect to the transformation parameters and that $\bR_{\theta\theta}$ is the cov(${\theta,\theta}$). The derivative can be obtained using analytical differentiation of $\mathrm{T}_{\theta}(\cdot)$ and the numerically calculated gradient of the moving image with respect to the coordinates. {\color{black} The gradient of the image with respect to coordinates is computed on the image lightly smoothed with gaussian kernel to alleviate gradient estimation noise. } We approximate $\bR_{\theta\theta}$ with the aid of Hessian $\bH$ of the similarity metric as $\bR_{\theta\theta} \propto \bH^{-1}_M$. A detailed discussion on the validity and applicability of such an approximation can be found in~\cite{Taron2009}. The Hessian $\bH$ is often available from the optimization routine. If not, its numerical approximation can be easily constructed. {\color{black} If Hessian $\bH$ is not available from the optimization routine or is too costly to compute, it is possible to replace it with the second order derivatives of the log-likelihood function resulting from model~\eqref{eqn:generative_model}. } We set $\bR_{\theta\theta} = \bH^{-1}_M$ and $\sigma^2_{\xi} = 1$ in our implementation. We have experimented with different settings of $\sigma^2_{\xi}$ ranging from 0.1 to 100 and found that the performance of the algorithm is not at all sensitive to the value of $\sigma^2_{\xi}$. To initialize the parameter $\mu_{\theta}$ at a given resolution scale we use its estimate obtained from the registration at the previous resolution scale.

Once utilities are calculated for each voxel we can compute the VSPF in accordance with~\eqref{eqn:pspfSolution}. As was mentioned in the previous section we keep the value of the free parameter $A$ minimal to diversify voxel selections and sufficient so that $\lambda^*$ exists such that the constraint equation ${\mathcal{P}^*}^T \bC = C_{\ave}$ can be satisfied. Furthermore, we compute the value of Lagrange multiplier $\lambda^*$ by finding the root of equation ${\mathcal{P}^*}^T \bC = C_{\ave}$ via bisection search (using the result stated in Proposition~\ref{prop:2}). In the constraint equation (see~\eqref{eqn:optProblem}), $C_{\ave}$ is the total average cost of processing the selected voxels. It is specified by the user e.g. in the form $C_{\ave} = c N$ where $c$ is the user specified average proportion of voxels being selected by the algorithm (the sampling rate). The high-level outline of the registration algorithm employing proposed voxel sampling algorithm is shown in Fig.~\ref{fig:alg_proposed}.

Finally, the last parameter of the algorithm is $P_{\mathrm{h}}$, the highest probability that could be assigned to a voxel. In the first approach that we explore, for every scale we simply adjust $P_{\mathrm{h}}$ heuristically such that more image exploration is induced in lower resolution scales while more emphasis is put on selecting voxels with higher utility in higher resolution scales. We found empirically, via experiments with the Brainweb~\cite{Kwan1999} dataset, that there exist simple heuristic schedules according to which $P_{\mathrm{h}}$ is set proportional to $C_{\ave} / N$ that produce universally good results. We present these schedules in more detail in Section~\ref{sec:Numerical_Experiments}.

The second approach that we propose is to learn $P_{\mathrm{h}}$ using a small dataset representative of the particular problem context. In the application domain,  the specifics of the particular registration problem often affect the choice of similarity metric, optimization strategy and interpolation scheme. This implies that at least some training information in the form of the small set of exemplar image pairs from the problem-specific modalities using certain acquisition and post-processing protocols should be available to the registration algorithm designer to guide the algorithm development.

Based on the assumption that we have a training data set and the gold standard registration parameters for the image pairs in this dataset we formulate the empirical learning criterion $Q^r(\Ph^r)$ for each resolution level $r$. We define the empirical target registration error (ETRE) as the average over $V$ image pairs in the training dataset and $U$ Monte-Carlo trials:
\begin{align}
Q^r(\Ph^r) = \frac{1}{V}\frac{1}{U} \sum_{v=1}^{V} \sum_{u=1}^U \| \bX_v - \widehat\bX^r_{u,v}(\Ph^r) \|_2^2.
\end{align}
Here $\bX_v$ is the set of transformed coordinates obtained using gold standard registration parameters for image pair indexed by $v$ and $\widehat\bX^r_{u,v}(\Ph^r)$ is the set of transformed coordinates for image pair $v$ and Monte-Carlo trial $u$ found using the empirical estimate of the registration parameters obtained via the optimization of the similarity metric at resolution scale $r$ using the proposed voxel sampling algorithm with a given value of $\Ph^r$. As the voxel sampling algorithm is randomized, some degree of Monte-Carlo averaging could be beneficial if $V$ is relatively small ($3\ldots5$ images). Thus we repeat the registration procedure for the same candidate value $\Ph^r$, level $r$ and image pair $v$ $U$ times and calculate $\widehat\bX^r_{u,v}(\Ph^r)$ based on the new registration parameter estimate each time.

We propose to learn the value of $\Ph^r$ by minimizing the ETRE $Q^r(\Ph^r)$:
\begin{align}
\widehat \Ph^r = \arg\min_{\Ph^r \in [0; 1]} Q^r(\Ph^r).
\end{align}
The function $Q^r(\cdot)$ is generally extremely irregular and non-smooth, because of the possible registration failures and because of complex dependence of the ETRE on the value of $\Ph^r$. At the same time, the domain of this function is well defined and restricted. Thus any optimizer capable of performing global or quasi-global search on a restricted interval using only the objective function values will suffice to solve this problem. In order to find $\widehat \Ph^r$ we propose using either the particle swarm optimization (PSO)~\cite{Kennedy1995} or the exhaustive grid search with grid step 0.01. Our algorithm proceeds by finding $\widehat\Ph^K$, the value for the scale with the lowest resolution. The algorithm then uses the identified value of $\widehat\Ph^K$ to find the estimate of the registration parameters at resolution level $K-1$. This procedure iterates until the values of highest probabilities for all resolution levels $K, K-1, \ldots, 1$ are found.

\section{Experiments with the RIRE Vanderbilt Dataset} \label{sec:Numerical_Experiments}

\subsection{RIRE Dataset}

To test the proposed algorithm we made use of the real clinical data available in the popular, publicly available RIRE Vanderbilt dataset~\cite{Fitzpatrick1998}. The performance of algorithms was evaluated by registering 3D volumes corresponding to CT images to geometrically corrected MR images. MR image set included images acquired using T1, T2 and PD acquisition protocols. The total number of different image pairs used was 19. Those pairs were taken from patients 001, 002, 003, 004, 005, 006, 007 for which geometrically corrected images are available. Patients 003 and 006 did not have geometrically corrected PD and MR-T1 images respectively. According to the data exchange protocol established by the RIRE Vanderbilt project, registration results obtained via algorithms under the test were uploaded to the RIRE Vanderbilt web-site. Algorithm evaluation results were calculated by the RIRE Vanderbilt remote computer using the gold standard transformation not available to us and published on their web-site in the form of tables containing target registration errors (TREs) calculated over 6 to 10 volumes of interest (VOIs) for each image pair. For patient 000 geometrically corrected MR-T1, MR-T2, MR-PD images and corresponding CT image are available along with the set of transformed coordinates obtained using gold standard registration parameters. Three image pairs from patient 000 were used to learn the values of $\Ph$ in one of the versions of the proposed algorithm according to the procedure described in Section~\ref{ssec:setting_Ph}.

\subsection{Experimental Setup}

\begin{figure*}[t]
\centering
    \subfigure[Coarse resolution level]{
    \label{fig:Ph_schedules:level2} %% label for second subfigure
    \includegraphics[width = \columnwidth]{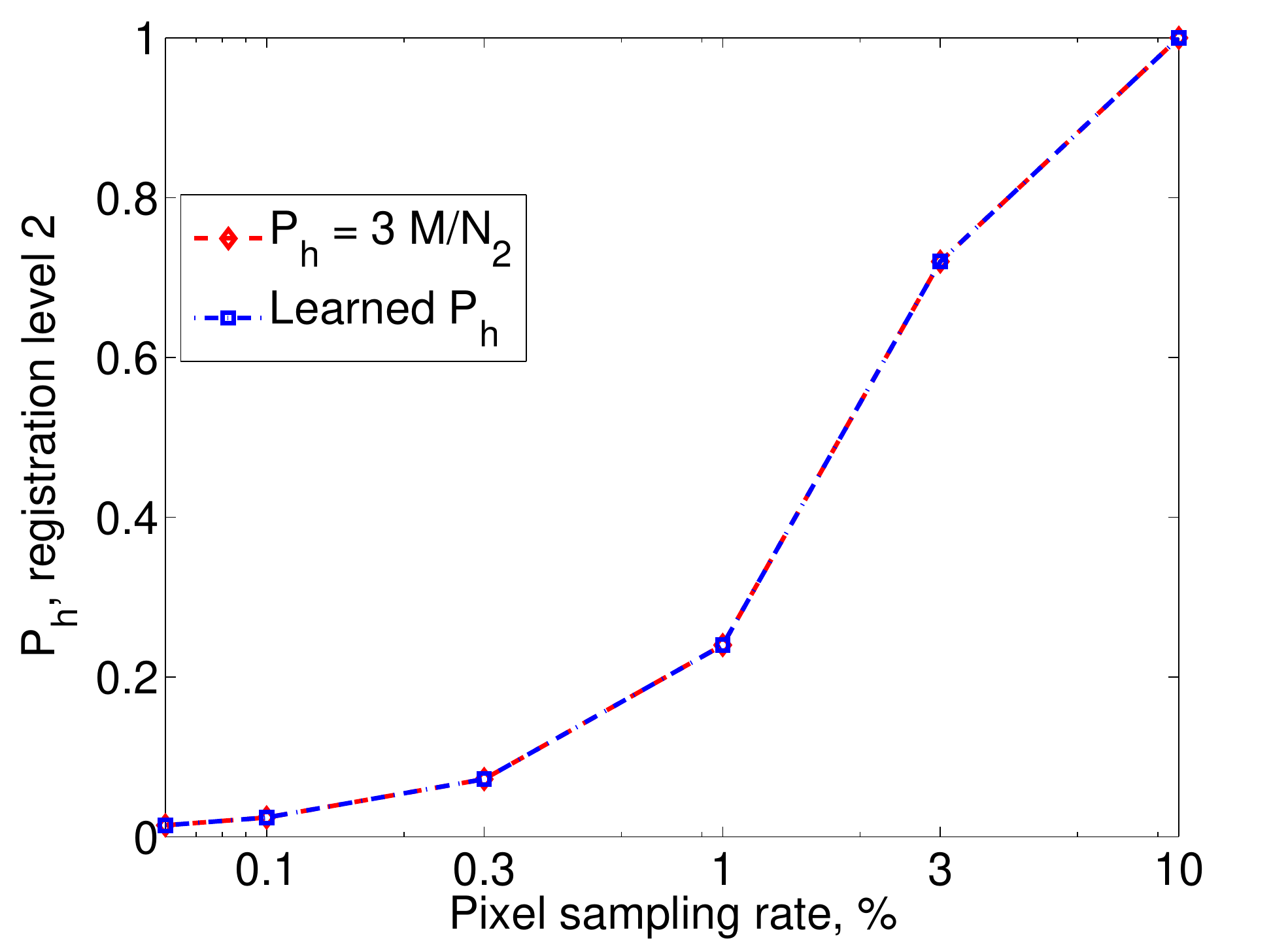}}
    \subfigure[Fine resolution level]{
    \label{fig:Ph_schedules:level1} %% label for second subfigure
    \includegraphics[width = \columnwidth]{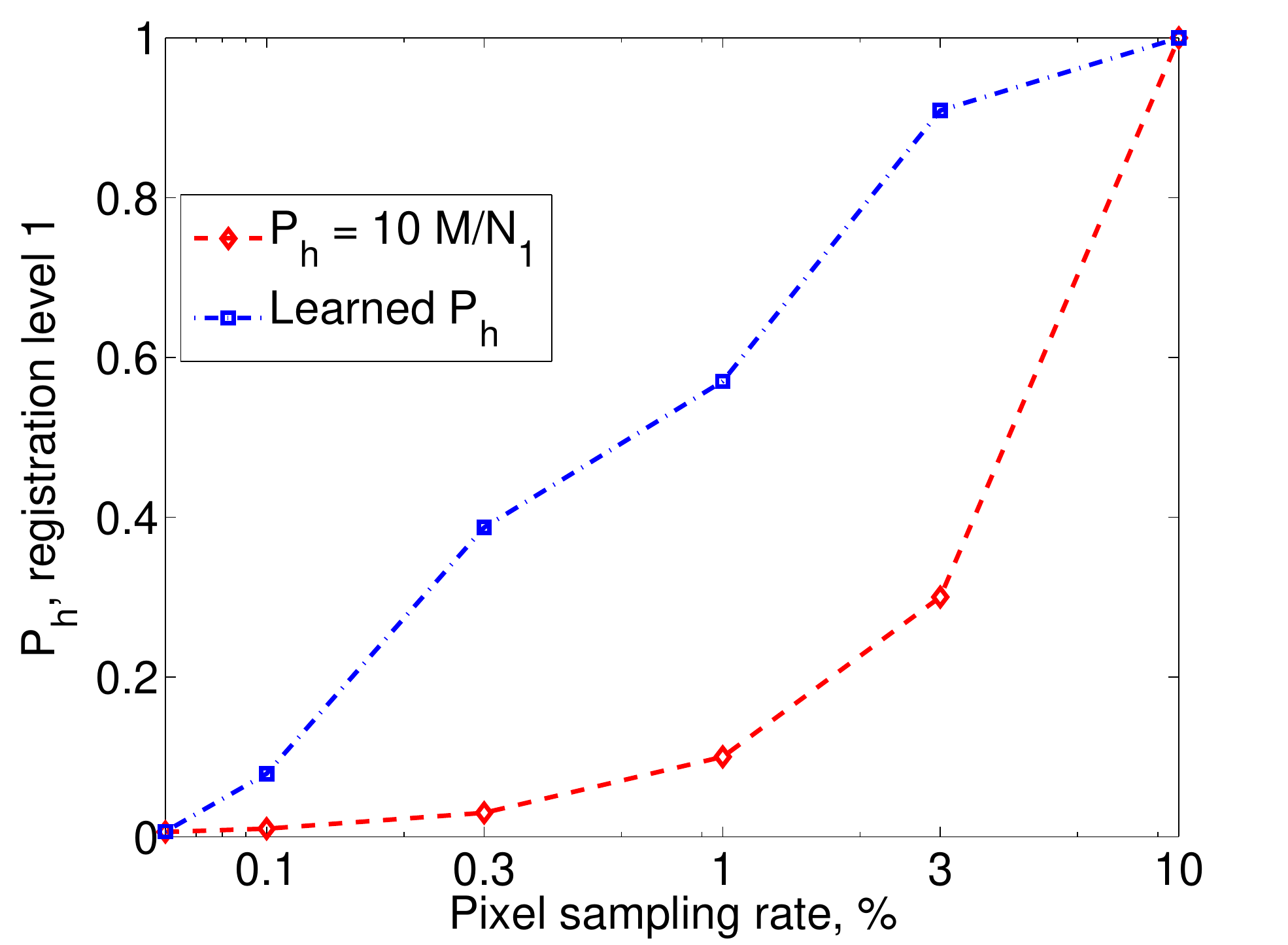}}
\caption{Heuristic and learned schedules of $\Ph$ for the proposed algorithm. {\color{black} Note that abscissa is in the log scale. } }
\label{fig:Ph} %% label for entire figure*
\vspace{-0.5cm}
\end{figure*}

All images were first resampled to a common 1mm grid using bicubic interpolation. We used 2-scale registration based on the low-pass filtered and downsampled image pyramid containing two resolution scales. Resolution level number two had grid spacing 2 mm along each axis and resolution level number one had grid spacing 1mm along each axis. The estimate of the registration parameters obtained at a lower resolution level was used as a starting point for the registration at the next higher resolution level; level 2 had all its parameters initialized to zero values. We concentrated on recovering 6 rigid registration parameters (3 translations and 3 rotations) using the NMI similarity metric~\cite{Studholme99}. Histogram for the evaluation of the similarity metric was calculated using the partial volume approach with Hanning windowed sinc kernel function~\cite{Lu2008}. The similarity metric was optimized using the trust region Gauss-Newton approach~\cite{Brooks2008}.

We evaluated the performance of the following voxel sampling approaches. The uniformly random sampling (\textbf{URS}) technique consists of randomly selecting voxels with equal probabilities at every iteration~\cite{Viola1997}. At a given resolution level $r$ all voxels have equal probability of being selected, $M / N_r$ if $M < N_r$ and $1$ if $M \geq N_r$; the average number of selected voxels is thus equal to $\min(M, N_r)$ at each resolution level. Note that we used equal number of selected voxels for all resolution scales. As a variation  of this scheme, we also considered the uniformly random sampling that samples voxels once per scale and keeps them fixed during iterations (\textbf{fURS})~\cite{Mattes2003}. Gradient magnitude sampling (\textbf{GMS}), a slight modification of gradient based subsampling originally proposed by Subuncu and Ramadge~\cite{Sabuncu2004}, consists of calculating the spatial gradient magnitude $\|\nabla V_i\|_2 = \sqrt{(\partial V_i/\partial x_i)^2 + (\partial V_i/\partial y_i)^2 + (\partial V_i/\partial z_i)^2}$ and sampling voxels at every optimization iteration according to the probabilities proportional to it, where the proportionality coefficient is chosen so that the average number of voxels selected at every resolution scale is equal to $M$. The extention of the previous approach, \textbf{GMS+URS},~\cite{Oreshkin2012}, samples voxels from the VSPF defined as a convex combination of VSPFs generated via GMS and URS approaches. The mixing parameter is optimized using a small training dataset (in our case this is the dataset obtained from patient 000).  For completeness we also present results for the gradient magnitude (\textbf{GM}) approach that deterministically selects a portion of voxels with highest gradient magnitudes once per scale. The proposed method consists of evaluating voxel utilities and assigning voxel probabilities based on these utilities. We evaluated two versions of the proposed algorithm. For the first option (\textbf{ProposedH}) we set $\Ph$ to $\min(1, 3 M / N_2)$ for the resolution level 2 (coarse resolution) and $\min(1, 10 M / N_1)$ for the resolution level 1 (fine resolution). These are the simple heuristic schedules (see Section~\ref{ssec:setting_Ph}). For the second option (\textbf{ProposedL}) we learn the value of $\Ph$ for every scale and voxel sampling rate using methodology described in Section~\ref{ssec:setting_Ph} with the aid of the training dataset consisting of the image pairs taken from patient 000. {\color{black} For completeness, we also present the \textbf{ProposedD} method, which is based on the thresholding of the proposed VSPF (ProposedD deterministically selects a portion of voxels with highest voxel utilities). We evaluate these algorithms for the following values of voxel sampling rates (given in \%): $M/N \in \{0.06, 0.1, 0.3, 1, 3, 10 \}$ (sampling rate is calculated with respect to the image size at the highest resolution level, $N = N_1$). } For the \textbf{GMS+URS} approach we learned the values of the mixing parameter via the exhaustive grid search procedure with grid step 0.01 and obtained the following results: {\color{black} $\beta^2 = [0.98; 0.93; 0.94; 0.97; 0.97; 0.95]$, $\beta^1 = [0.33; 0.29; 0.25; 0.22; 0.09; 0.00]$. } The mixing parameter (denoted as $\beta^r$ in the original paper~\cite{Oreshkin2012}) is responsible for finding a good balance between the GMS and the URS approaches within the \textbf{GMS+URS} sampling scheme. Each value of $\beta^r$ presented above corresponds to one of the sampling rates used to evaluate the algorithms.

\subsection{Results} \label{ssec:Results}

Figure~\ref{fig:Ph} compares two schedules for setting the $\Ph$ parameter of the proposed algorithm. The heuristic schedule $\min(1, 3 M/N_2)$ for coarse resolution and $\min(1, 10 M/N_1)$ for high resolution level is contrasted with the schedule learned using small training dataset methodology presented in Section~\ref{ssec:setting_Ph}. {\color{black} Note that the curve corresponding to the heuristic schedule does not appear to be linear in the plot, because abscissa is in logarithmic scale.} It can be seen from Fig.~\ref{fig:Ph_schedules:level2} that for the coarse resolution level there is not much difference in the heuristically proposed and the learned schedules. On the other hand, Fig.~\ref{fig:Ph_schedules:level1} showing results for the higher resolution level reveals that, according to the learned schedule, $\Ph$ grows at a much steeper rate than $10 M/N_2$. This suggests that for the higher resolution level there is more variation in the utility of different voxels and there is a need for choosing higher values of $\Ph$ to enhance the stratification among voxel probabilities to boost selection frequencies for more informative voxels. Another interpretation of this result is that when the total average number of sampled voxels becomes large enough, we can allow the sampling scheme to focus almost exclusively on voxels with high utility values, because the exploratory power of this voxel subset becomes strong enough.

\begin{figure*}[t]
\centering
      \subfigure[ProposedL (ProposedL was visually similar to ProposedH)]{
      \includegraphics[width=0.35\textwidth, angle=90]{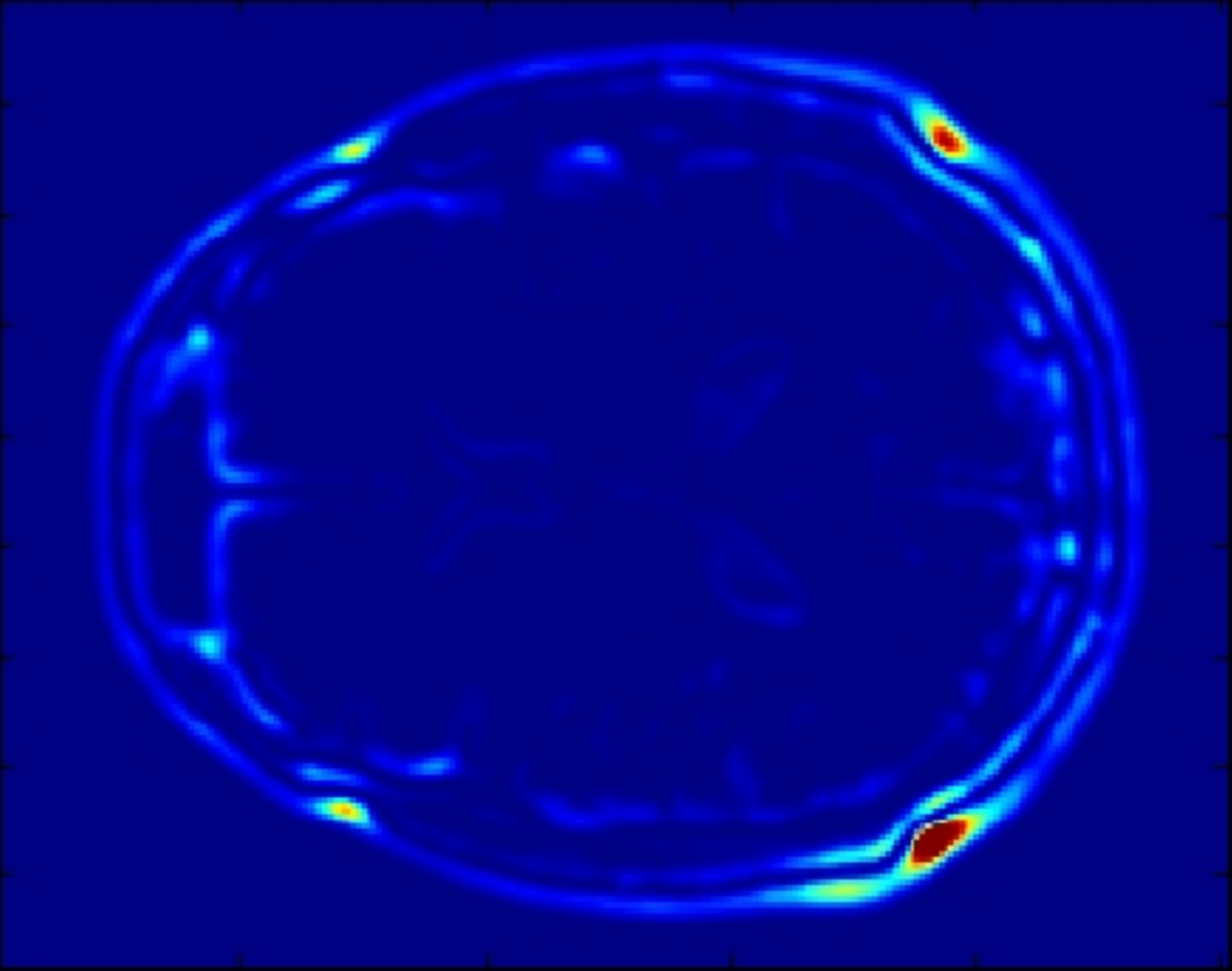}}
      \subfigure[ProposedD]{

      \includegraphics[width=0.35\textwidth, angle=90]{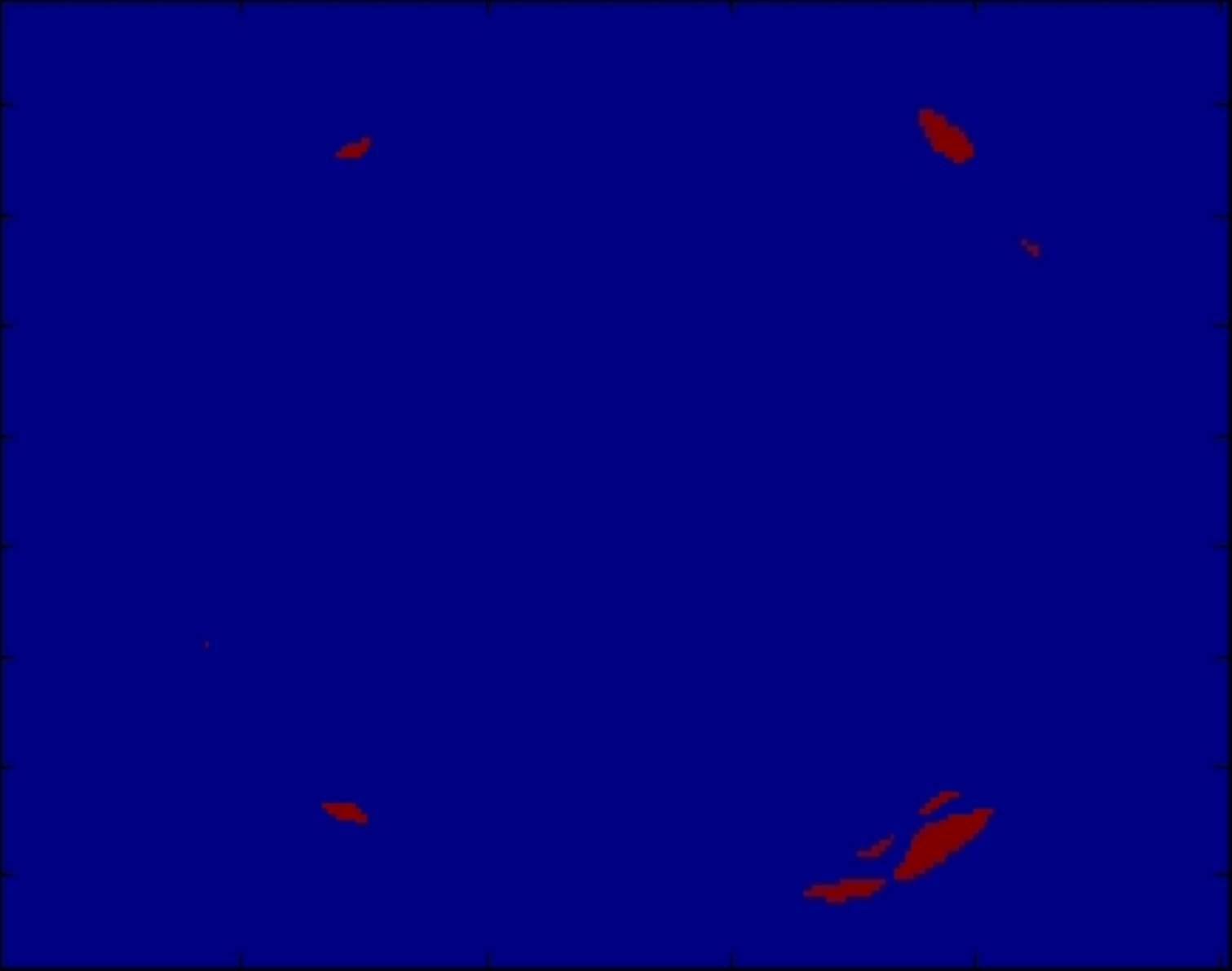}}
      \subfigure[GMS combined with URS, GMS+URS]{
      \includegraphics[width=0.35\textwidth, angle=90]{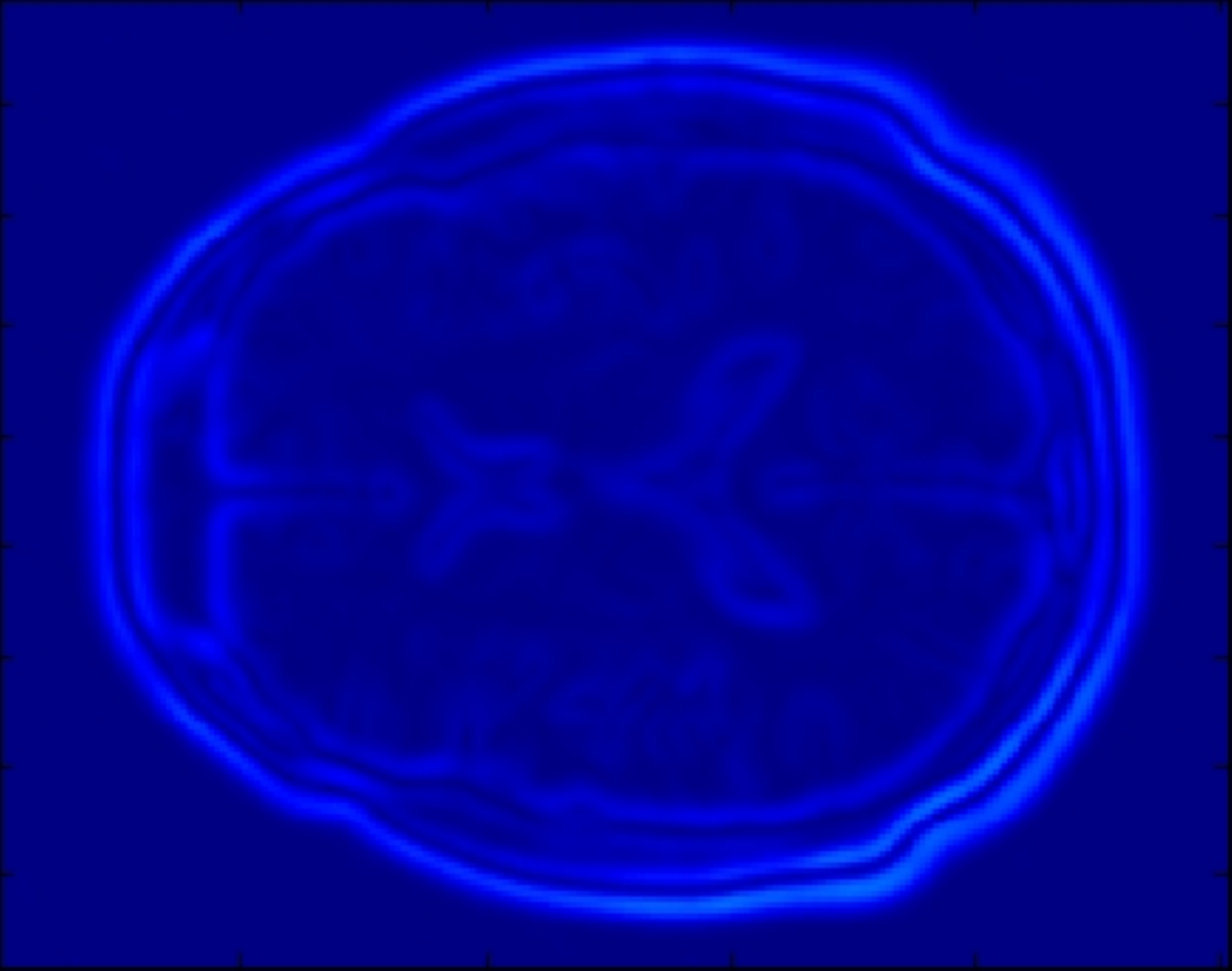}}
      \subfigure[gradient subsampling, GMS]{
      \includegraphics[width=0.35\textwidth, angle=90]{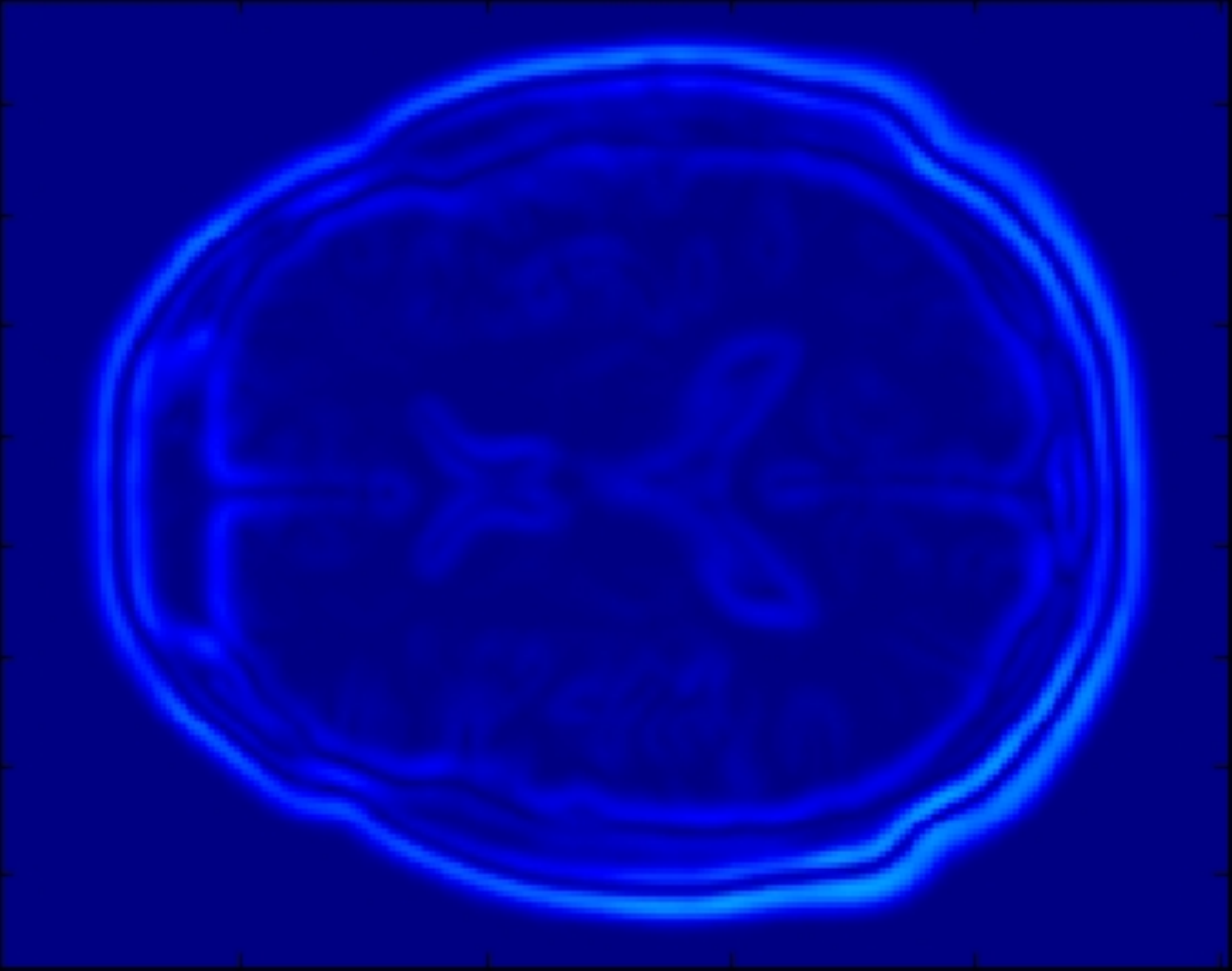} }
      \subfigure[gradient magnitude, GM]{
      \includegraphics[width=0.35\textwidth, angle=90]{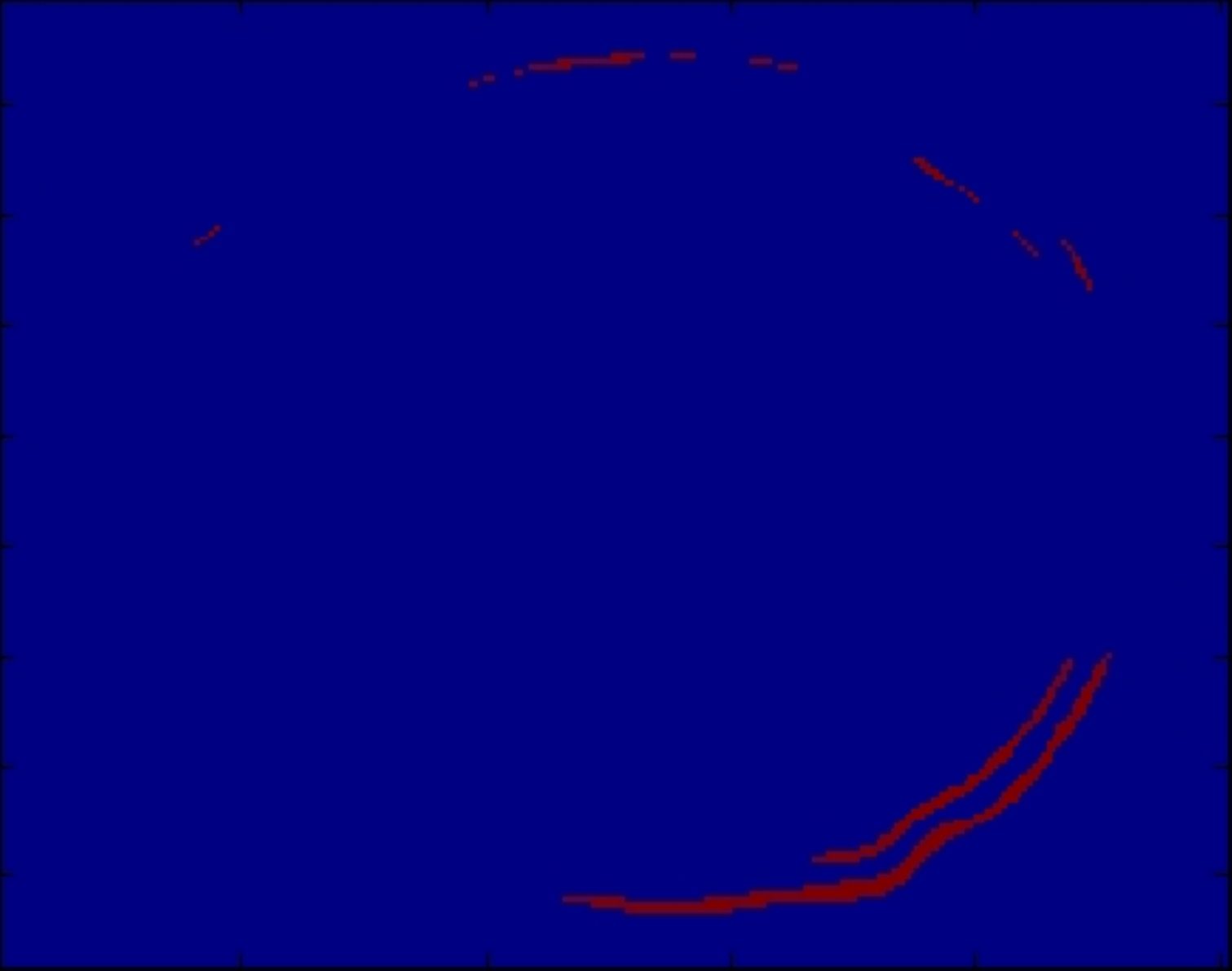} }
%\\
\caption{VSPFs for the highest resolution level shown in jet colour scheme. The jet color scheme uses colour assignment as follows: black  corresponds to zero ($p_i=0$); cyan, yellow and orange correspond to intermediate values between $0$ and $1$; red corresponds to ($p_i = 1$).}
\label{fig:VSPF_examples} %% label for entire figure*
\vspace{-0.5cm}
\end{figure*}

Qualitative results are presented in Figure~\ref{fig:VSPF_examples} showing an example of VSPFs obtained via the {\color{black} ProposedL, ProposedD, GMS+URS, GMS and GM } approaches (URS and fURS are not shown since their VSPFs are uniform and ProposedH was visually similar to ProposedL). The figure is obtained for $C_{\ave} = 0.01 N$, corresponding to 1\% of voxels being sampled on average. We can see that the proposed approach allows for the selection from a comparatively large variety of voxels (compared e.g. to the GM or {\color{black} ProposedD } approaches). We also see that there are voxels in the proposed VSPF that are assigned significantly higher probabilities than some other voxels. These voxels are deemed more informative by our sampling scheme. The GMS sampling scheme assigns probabilities proportional to voxel gradient magnitudes. The GMS+URS scheme tries to find the compromise between GMS and URS schemes by adding uniform probabilities to the GMS VSPF. The GM scheme selects a subset of voxels with highest gradient magnitudes and assigns them probabilities equal to 1 (voxels are always selected). The selection of voxels produced by the GM scheme is not rich as they are mostly concentrated along the most prominent spatial gradient structures. {\color{black} Similar comments apply to the ProposedD mask that selects a subset of voxels having most prominent utility values. } Moreover, note that the proposed approach favours more peripheral voxels useful for angular alignment, as opposed to the GM, GMS and GMS+URS approaches that do not account for parameter uncertainty and concentrate only on image gradient details. We argue that this characteristic, along with the ability to diversify voxel selections brings forward an improved registration robustness and accuracy. On the other hand, large areas populated with uninformative voxels have zero VSPF values. Thus, by concentrating image exploration only on potentially informative voxels, the proposed approach improves the accuracy over the GMS+URS, URS and fURS approaches.

{\color{black}

\begin{figure*}[t]
\centering
    \subfigure[Failure rate]{
    \label{fig:performancePlots:FailureRate} %% label for second subfigure
    \includegraphics[width = \columnwidth]{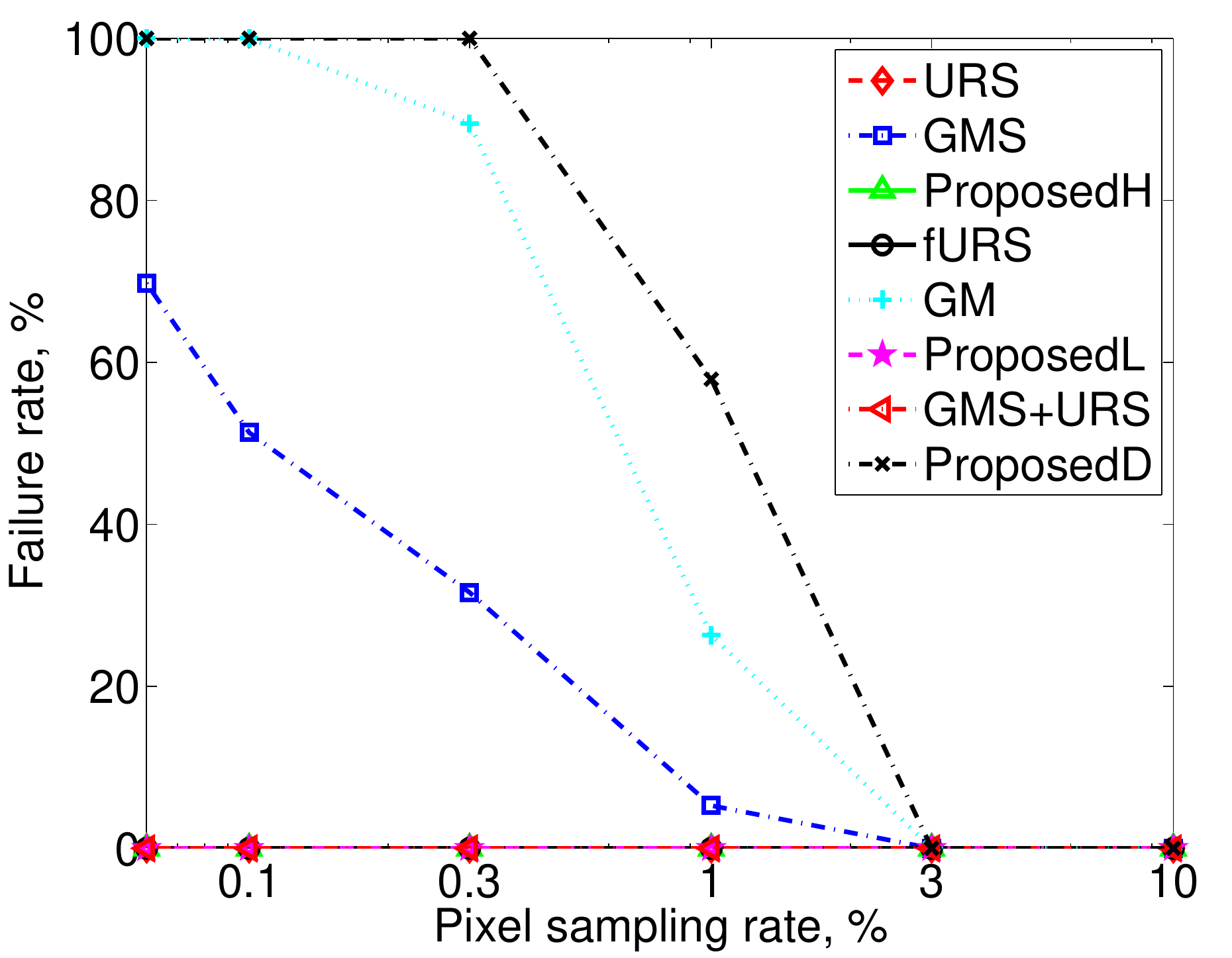}}
    \subfigure[mTRE]{
    \label{fig:performancePlots:mTRE} %% label for second subfigure
    \includegraphics[width = \columnwidth]{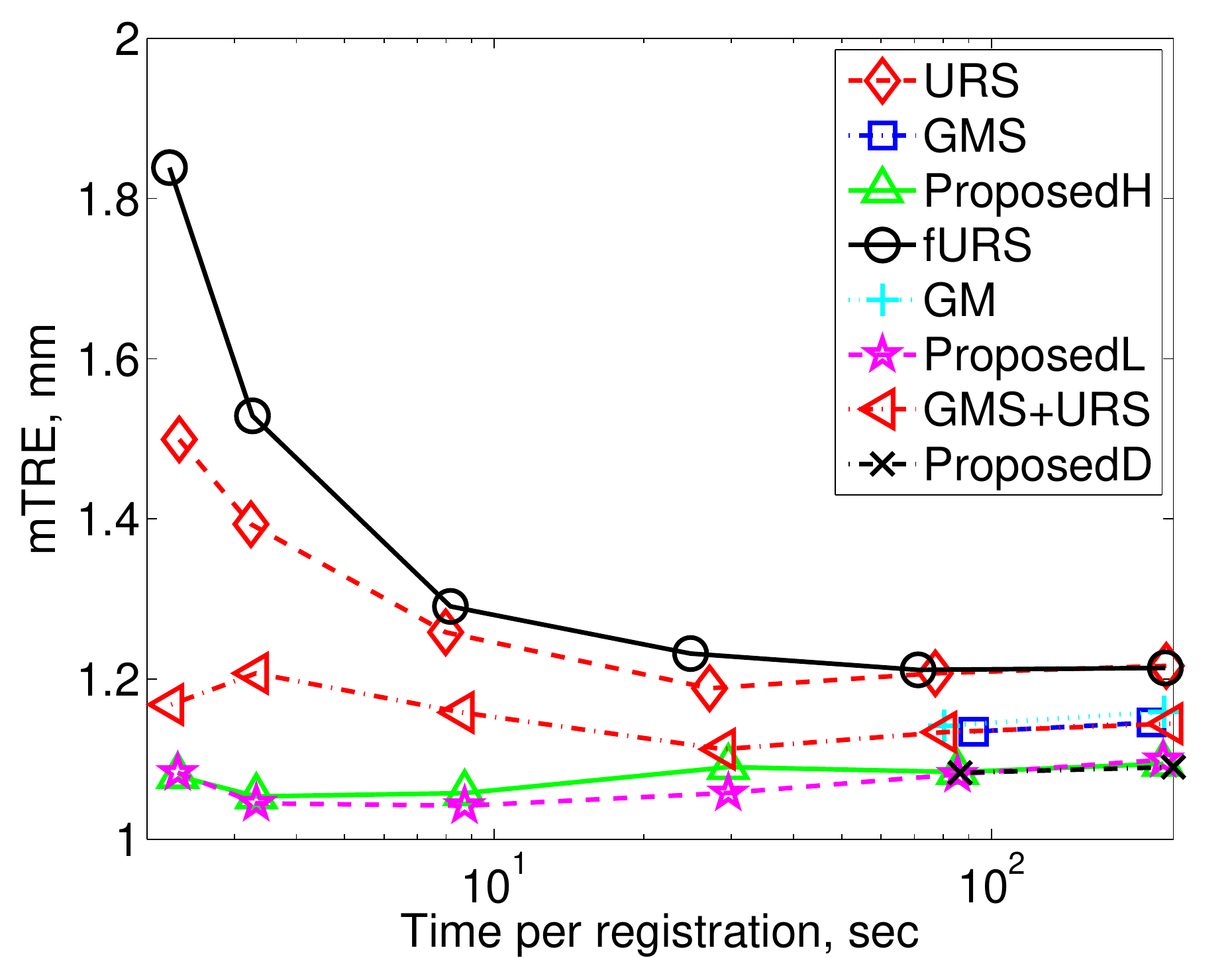}}
\caption{Failure rate (a) and mTRE (b) for different voxel selection mechanisms: gradient magnitude (GM), gradient magnitude sampling (GMS), proposed with learned $\Ph$ (ProposedL) and heuristic $\Ph$ (ProposedH), ProposedD corresponding to the thresholded proposed VSPF, uniformly random sampling (URS), uniformly random sampling with fixed subsets (fURS), convex combination of GMS and URS (GMS+URS). Note that the proposed approach consistently outperforms in terms of robustness and accuracy.}
\label{fig:performancePlots} %% label for entire figure*
\vspace{-0.5cm}
\end{figure*}

}

Figure~\ref{fig:performancePlots:FailureRate} shows registration failure rate for different voxel selection mechanisms as a function of the sampling rate shown in the log scale. We define a registration failure as any case with error exceeding 10mm in any of the VOIs. The $x$ axis of this plot shows voxel sampling rates $\{ 0.06, 0.1, 0.3, 1, 3, 10 \}$. We can see that both variants of the proposed approach employing sampling from the VSPF consistently outperform other approaches in terms of robustness, because both methods employ heavy exploration of the voxel values by continuous sampling combined with emphasizing more important voxels.

Figure~\ref{fig:performancePlots:mTRE} shows the mTRE (mean TRE) as a function of the average time spent per registration (note that the x-axis is in the log scale). Recall that RIRE provides us with 6 to 10 TRE values for each image pair. We compute the mTRE as the mean TRE of all the successful (non-failed) registrations corresponding to a particular experimental case defined by the selection of voxel sampling scheme and voxel sampling rate. For objective comparison we show only those points that correspond to zero failure rate, because we deem this scenario as the one that has most practical importance. Besides, this helps avoiding the situation where a method demonstrates anomalously low mTRE, because it fails many difficult cases. The markers in Figures~\ref{fig:performancePlots:mTRE} and~\ref{fig:performancePlots:FailureRate} correspond to the same instances of the sampling rate. It can be seen that the average time per registration is roughly proportional to the sampling rate (for the proposed method the calculation of the VSPF induces only a small overhead). The mTRE is minimal for the proposed approach with learned $\Ph$ (ProposedL) compared to other methods. {\color{black} The ProposedD method has the best performance for sampling rates 3 and 10\%, however, it simply does not work for lower sampling rates because of failures. } Same applies to the GMS and and the GM approaches, both having only two points shown in the plot because of failures. The proposed approach retains high level of accuracy and robustness even with low voxel sampling rates. Moreover, the proposed approach with the heuristic setting of $\Ph$ (ProposedH) is not much worse than the proposed approach with learned $\Ph$ (ProposedL). This shows relative insensitivity of the proposed approach to the changes in parameter $\Ph$. The properties of the proposed sampling approach allow to significantly reduce computational time and simultaneously guarantee high accuracy and robustness in a practical system. Such results support our conjecture that image exploration induced by VSPF based voxel sampling maintains the robustness due to continued image exploration and improves registration accuracy by selecting more informative voxels. Overall, the proposed technique at $0.06\%$ sampling rate is better than other techniques at $10\%$ sampling rate, maintaining $0\%$ failure rate and around $1$ mm mTRE accuracy. This amounts to more than hundredfold acceleration for the registration procedure. None of the alternative techniques achieve this level of performance.

{ \color{black}

\section{Discussion of Findings} \label{sec:dicussion_of_findings}

We have demonstrated the promise of the proposed theoretical methodology based on the experiments with the 3D RIRE Vanderbilt dataset containing real clinical data. The main finding of the paper is that with the aid of a carefully designed VSPF (in our case relying on the principle of the Bayesian registration parameter uncertainty) we can maintain the appropriate levels of accuracy and robustness of a registration procedure at very low voxel sampling rates.

At the same time, our validation methodology has a number of limitations and calls for further investigation and analysis of the properties of the proposed theoretical framework. We conducted our experiments in the context of rigid registration. In some contexts rigid registration is appropriate even during image guided interventions~\cite{Roche01}. However, some other contexts may call for the use of non-rigid registration approaches (e.g. in the context of image guided neurosurgery, open craniotomies cause non-rigid intra-operative brain movements also known as brain shift~\cite{Arbel01}). The proposed theoretical framework does not pose any principal limitations on the size or properties of the registration parameter set. At the same time, with the current validation {\color{black} setup} we cannot firmly state if the proposed methodology is directly applicable to the registration problems with non-rigid parameterization. Intuitively, however, we may expect even better performance gains for the non-rigid registration problems. The reason for this is that the data (voxel) support for each of the registration parameters is highly spatially localized. The effects of differing voxel utilities (abilities to reduce registration error) are likely to be even more pronounced in this setting. This should result in increased gain from a smart sampling strategy and may call for further extensions. For example, including parameters into the sampling scheme and deciding on a subset of parameters to update at a given iteration. Often, many parameters in complex non-rigid registration problems converge early and do not require frequent updates during the rest of the procedure.

Another limitation of the validation methodology is that the evaluation of the proposed method was performed on the RIRE Vanderbilt dataset, which does not necessarily target real-time applications. Real-time applications often use noisy images, resulting e.g. from ultrasound or fluoro modalities~\cite{Sakai06,Roche01}. Based on the current validation scheme it is not clear if the proposed method will provide the same performance gain in the case of the low quality images. The study of the proposed method in this context is thus a necessary important future research task. At the same time, we would like to note that our VSPF calculation methodology requires only one of the images to compute the sampling probabilities. In the common real-time applications outlined above one of the images: MRI, CT or another high quality image, is typically available before the application of the registration procedure. In this context the high quality image can be defined as a moving image and used to compute the VSPF off-line. This scenario is very close to our current validation {\color{black} setup} in terms of the images used to compute the VSPF. Our validation setup has its limitations, but we believe that it is rigorous enough to demonstrate the promise of the proposed theoretical framework and to approximate some of the realistic real-time applications.

Another important finding is that from our experiments (see Fig~\ref{fig:performancePlots:mTRE}) it follows that, at least for the rigid registration problem, the calculation of the VSPF does not induce any significant overhead. Recall that we compute the VSPF only once per registration scale and do not recompute it at every iteration. With this scheme we achieve significant speed-up in the registration procedure and keep robustness and accuracy of the method at very high levels. Although the complexity of computing the VSPF scales linearly with the number of registration parameters, so does the complexity of computing the similarity metric and its derivatives. Thus we have a reason to believe that the implementation of the proposed methodology in the context of non-rigid registration problems will not result in any disproportional overhead. Furthermore, for many problems the high-quality image (e.g. diagnostic MRI) is available well before the registration procedure must be executed. In this context we may further reduce the overhead by computing the VSPF off-line and then just reusing it during the operation. Moreover, we can imagine much more complicated sampling schemes being implementable in such scenario. For example, for a non-rigid problem we could define its own VSPF for every registration parameter and compute it off-line. During the application of the registration procedure we would sample optimal voxels for each parameter individually and then combine all of them into the common set to evaluate the similarity metric.

Our sampling approach uses different pixel subsets at every optimization iteration. This results in the continued exploration of the images being registered, but also induces noise in the approximations of the cost function and its derivatives. In the context of the Gauss-Newton optimization used in our paper, changing voxel selections over iterations does not pose any problem. It should be noted that although we presented the concepts of VSPF and continued voxel sampling within the Gauss-Newton optimization framework, they have a much wider applicability. For example, the global CMAES optimizer~\cite{Hansen2009} widely used for complex optimization problems (including the non-rigid registration) is capable of handling the noise in cost function values. Different variations of gradient descent algorithms are not sensitive to noise in derivatives as well, especially when combined with stochastic optimization approaches~\cite{ccPOL92}. Thus the proposed framework can be used with these optimization approaches directly. At the same time, the proposed framework is not directly applicable to the quasi-Newton (non-zero memory) optimizers where an estimate of the Hessian of the cost function is built up iteratively. It may well be that the proposed sampling strategy could be adapted for the use with such optimizers (e.g. via proper use of common random numbers techniques~\cite{Kahn53}). However, investigating this matter is beyond the scope of this paper.

Finally, the current approach is formulated for the direct registration methods using only voxel values. It would be interesting to look for the extensions of this method in the context of feature based registration. To directly extend the proposed method in this context one could formulate a Bayesian model for the analysis of registration parameter uncertainty as a function of selected feature set. The next step in this direction would be to define and optimize the feature sampling probability field based on the uncertainty analysis.

}

\section{Conclusions} \label{sec:Conclusions}
We have presented a new uncertainty based multi-scale voxel selection framework for the context of multi-modal medical image registration. At each resolution scale, a voxel sampling probability field (VSPF) is optimized based on the transformation parameter uncertainties derived from the Bayesian model and the optimal voxel subsets are then sampled from the VSPF. The proposed sampling scheme exhibits the advantages of both random sampling techniques, which are robust but not necessarily accurate, and fixed sampling techniques, which are accurate but not necessarily robust. Specifically, the proposed sampling strategy focuses on both informative image details that have potential to contribute towards reducing registration uncertainty whilst simultaneously maintaining the variability and exploratory power of the sampled voxel set. The experimental results on the 3D, multimodal, publicly available RIRE dataset demonstrate that our approach yields significant improvements in registration performance, compared to other voxel selection techniques such as random voxel sampling and gradient sampling, particularly at aggressive sampling levels (e.g. less than $1\%$ of voxels). Overall, the proposed technique at $0.06\%$ sampling rate was better than other tested techniques at $10\%$ sampling rate, maintaining $0\%$ failure rate and around $1$ mm mTRE accuracy. This amounts to more than hundredfold reduction in the number of voxels used for similarity metric calculations, which translates to similar accelerations for the registration procedure. This leads to minimization of the computational time of the registration task while also minimizing the potential penalty in terms of accuracy and robustness (i.e. failure rate). The approach is general and can provide an efficient solution applicable to a wide array of important, time-sensitive medical imaging registration problems, such as is required in image guided interventions. Future work will focus on experimental validation of the registration scheme in various interventional contexts.

\appendices

{\color{black}

\section{Analysis of Bayes Uncertainty: Technical Details} \label{app:analysis_of_bayes_uncertainty}

First, we introduce some probabilistic notation. A real valued vector random variable $X : \Omega \rightarrow \mathbb{R}^d$ is defined on
the probability space $(\Omega, \mathcal{F}, P)$ equipped with sample space $\Omega$, $\sigma$-algebra of events $\mathcal{F}$ and probability measure $P$. The mathematical expectation is then defined as the following Lebesgue integral:
\begin{align}
\bbE\{X\} = \int_{\Omega} X(\omega) P(\mathrm{d}\omega).
\end{align}
For two random vectors $Y$ and $X$ we also define the covariance matrix $\bR_{YX}=\cov(X, Y)$ as
\begin{align}
\cov(X, Y) = \bbE\{(X-\bbE\{X\})(Y-\bbE\{Y\})^T\},
\end{align}
and variance $\var(X) = \cov(X, X)$.

Under our assumptions $\mathcal{A}_1$, $\mathcal{A}_2$ and by the definition of VSPF we have three primary random vectors in our paper: the vector of registration parameters $\theta$, the sequence of modeling noises $\xi_i, i=1\ldots N$ and the vector of decision variables $\mathcal{D}$. In what follows all the expectations are taken with respect to the joint probability measure induced by the collection of these three random vectors unless specifically stated otherwise or indicated by the conditional expectation $\bbE\{X|Y\}$ defined as:
\begin{align}
\bbE\{X|Y\} = \int_{B \subseteq \Omega : Y=Y(\omega)} X(\omega) P(\mathrm{d}\omega).
\end{align}

According to the model \eqref{eqn:generative_model} and under assumptions $\mathcal{A}_1$ and $\mathcal{A}_2$, $\moving_{i}(\mathrm{T}_{\theta}(\bx))$ has a Gaussian distribution with mean $\moving_{i}(\mathrm{T}_{\mu_\theta}(\bx))$ and variance $\bg_{i}^T\bR_{\theta\theta} \bg_{i} + \sigma^2_{\xi}$.
If we pose the problem of optimally estimating the parameters $\theta$ from the generative model~\eqref{eqn:generative_model} then the Bayesian estimate of parameters $\theta$ is simply
\begin{align}
\widehat\theta = \mu_{\theta} + \bR_{\theta
\moving} \bR_{\moving \moving}^{-1}
(\moving - \mu_{\moving}).
\end{align}
where $\mu_{\moving} = \bbE\{ \moving \}$, $\bR_{\moving \moving}
=\var(\moving)$, $\bR_{\theta \moving} = \cov(\theta,  \moving)$. On the other hand, the optimal Bayesian estimate $\widehat\theta_{\mathcal{D}}$ obtained using a given realization of random voxel selection decisions vector $\mathcal{D}$ and associated vector of selected voxels $\moving_{\mathcal{D}}$ is
\begin{align}
\widehat\theta_{\mathcal{D}} = \mu_{\theta} + \bR_{\theta
\moving_{\mathcal{D}}} \bR_{\moving_{\mathcal{D}} \moving_{\mathcal{D}}}^{-1}
(\moving_{\mathcal{D}} - \mu_{\moving_{\mathcal{D}}}).
\end{align}
where we denote $\mu_{\moving_{\mathcal{D}}} = \bbE\{ \moving_{\mathcal{D}} | \mathcal{D} \}$, $\bR_{\moving_{\mathcal{D}} \moving_{\mathcal{D}}}
=\cov(\moving_{\mathcal{D}}, \moving_{\mathcal{D}} | \mathcal{D})$, $\bR_{\theta \moving_{\mathcal{D}}} = \cov(\theta,  \moving_{\mathcal{D}} | \mathcal{D})$. The estimation error for $\widehat\theta_{\mathcal{D}}$ can be calculated using the law of total variance:
\begin{align}
\var(\theta - \widehat\theta_{\mathcal{D}}) = \bbE\{ \var(\theta -
\widehat\theta_{\mathcal{D}} | \mathcal{D})  \} + \var(\bbE\{ \theta
- \widehat\theta_{\mathcal{D}} | \mathcal{D} \}).
\end{align}
Since $\bbE\{ \moving_{\mathcal{D}} - \mu_{\moving_{\mathcal{D}}}  |
\mathcal{D} \} = 0 $ implies $\bbE\{ \theta
- \widehat\theta_{\mathcal{D}} | \mathcal{D} \} = 0$ we have $\var(\theta - \widehat\theta_{\mathcal{D}}) = \bbE\{ \var(\theta -
\widehat\theta_{\mathcal{D}} | \mathcal{D})  \}$. For a fixed $\mathcal{D}$ we can calculate the conditional variance
as
\begin{align}
\var(\theta - \widehat\theta_{\mathcal{D}}| \mathcal{D}) =
\bR_{\theta \theta} - \bR_{\theta \moving_{\mathcal{D}}}
\bR_{\moving_{\mathcal{D}} \moving_{\mathcal{D}}}^{-1} \bR_{\theta
\moving_{\mathcal{D}}}^T.
\end{align}
Under the assumption $\mathcal{A}_3$ implying that the covariance matrix $\bR_{\moving_{\mathcal{D}} \moving_{\mathcal{D}}}$ is diagonal for any configuration of $\mathcal{D}$ the above expression can be simplified as follows:
\begin{align}
\var(\theta - \widehat\theta_{\mathcal{D}}| \mathcal{D}) =
\bR_{\theta \theta} - \sum_{i=1}^N d_i \bR_{\theta \moving_{i}} \bR_{\moving_{i}
\moving_{i}}^{-1} \bR_{\theta \moving_{i}}^T.
\end{align}
Finally, upon evaluating the expectation with respect to $\mathcal{D}$ we obtain expression~\eqref{eqn:theta_var}.

}

\section{Solution of Optimization Problem~\eqref{eqn:optProblem}} \label{app:OptimizationSolution}

After rearranging terms in~\eqref{eqn:pspfCost} we can write:
\begin{align}
{\mathcal{J}}(\mathcal{P}) &= -\sum_{i=1}^N p_i (U_i + \gamma_i - \eta_i + \lambda \bC) - \sum_{i=1}^N \Ph \eta_i + \lambda C_{\ave} \nonumber \\
&= -\mathcal{P}^T (\bU + \bm{\gamma} - \bm{\eta} + \lambda \bC) - \sum_{i=1}^N \Ph \eta_i + \lambda C_{\ave} \nonumber \\
&= -\|\mathcal{P} \|_2 \| \bU + \bm{\gamma} - \bm{\eta} + \lambda \bC \|_2 \cos \beta - \sum_{i=1}^N \Ph \eta_i + \lambda C_{\ave}.
\end{align}
Here in the last two expressions we use vector notation $\bU = [U_1, U_2, \ldots, U_N]^T$, $\bm{\gamma} = [\gamma_1, \gamma_2, \ldots, \gamma_N]^T$, $\bm{\eta} = [\eta_1, \eta_2, \ldots, \eta_N]^T$, $\bC = [C, C, \ldots, C]^T$ and $\beta$ is the angle between vectors $\mathcal{P}$ and $\bU + \bm{\gamma} - \bm{\eta} + \lambda \bC$.

{\color{black}
The Lagrange relaxation method~\cite{Everett63} allows us to simplify the original minimization problem by first solving the unconstrained problem $\mathcal{P}^* = \min_{\mathcal{P}} {\mathcal{J}}(\mathcal{P})$ while treating the Lagrange multipliers as constants and then identifying the values of the Lagrange multipliers via original constraints.

In the first step of Lagrange relaxation, the last expression is minimized when $\cos\beta = 1$, i.e. vectors $\mathcal{P}$ and $\bU + \bm{\gamma} - \bm{\eta} + \lambda C$ are collinear and proportional: $\mathcal{P}^* = A(\bU + \bm{\gamma} - \bm{\eta} + \lambda C)$ with proportionality constant $A > 0$.
}

This can be proven easily by considering some arbitrary $\mathcal{P}^\star$ with $\| \mathcal{P}^\star \|_2 = c, 0 < c < \infty$ for which, necessarily, $\cos \beta < 1$. A better solution, $\mathcal{P}^\maltese$, can always be found by picking $\mathcal{P}^\maltese = \frac{c}{\| \bU +
\bm{\gamma} - \bm{\eta} + \lambda C \|_2} (\bU + \bm{\gamma} - \bm{\eta} + \lambda C)$. Inserting $\mathcal{P}^\maltese$ into the
expression for ${\mathcal{J}}$ demonstrates that ${\mathcal{J}}(\mathcal{P}^\maltese)
< {\mathcal{J}}(\mathcal{P}^\star)$.

{\color{black}
We can thus replace the original problem by introducing the positive auxiliary variable $A$ and reformulating the problem as follows:
\begin{align} \label{eqn:optProblem_auxiliary}
\mathcal{P}^* &= \min_{\mathcal{P}, A} {\mathcal{J}}(\mathcal{P}) \nonumber \\
\text{such that \quad} p_i \geq
0, \quad p_i &\leq \Ph, \quad \sum_{i=1}^N p_i C = C_{\ave}, \nonumber\\
p_i &= A(U_i + \gamma_i - \eta_i + \lambda C).
\end{align}

In the second step of Lagrange relaxation we identify the Lagrange multipliers utilizing the constraint conditions derived from the original constraints: }
\begin{align}
\gamma_i \leq 0, p_i &\geq 0, \gamma_i p_i = 0 \\
\eta_i \leq 0, p_i &\leq \Ph, \eta_i (\Ph-p_i) = 0 \\
\mathcal{P}^T \bC &= C_{\ave}
\end{align}
First we deduce the values of $\bm{\gamma}$ and $\bm{\eta}$. If $0 \leq A(U_i + \lambda C) \leq \Ph$, constraints are satisfied
and thus $\gamma_i=0$, $\eta_i = 0$. If $A(U_i + \lambda C) < 0$,
constraint $p_i \leq \Ph$ is satisfied and $\eta_i = 0$, but
$\gamma_i\neq 0$. To satisfy condition $\gamma_i p_i = 0$, we set
$\gamma_i = -(U_i + \lambda C)$ resulting in $p_i = 0$. If $A(U_i
+ \lambda C) > \Ph$, constraint $p_i \geq 0$ is satisfied and
$\gamma_i=0$. To satisfy condition $\eta_i (\Ph-p_i)=0$, we set
$\eta_i = \Ph A^{-1} - (U_i + \lambda C)$ resulting in $p_i = \Ph$.

The resulting solution can thus be represented as
\begin{align} \label{eqn:technicalSolution}
p_i^* = \left\{  \begin{matrix} A(U_i + \lambda^* C) & \text{ if } & 0 \leq A(U_i + \lambda^* C) \leq \Ph  \\
\Ph & \text{ if } & A(U_i + \lambda^* C) > \Ph  \\
0 & & \text{otherwise}
\end{matrix} \right.
\end{align}
Here the value $\lambda^*$ of the remaining Lagrange multiplier can be found by solving the constraint equation ${\mathcal{P}^*}^T \bC = C_{\ave}$ for $\lambda^*$. {\color{black} The auxiliary variable $A$ should be determined by the minimization problem~\eqref{eqn:optProblem_auxiliary} and we will later show how this can be accomplished by analyzing the properties of the cost function. }

\section{Proof of proposition~\ref{prop:1}} \label{app:ProofOfProposition1}

\begin{IEEEproof}
We will now show that ${\mathcal{J}}(\mathcal{P}^*), \forall A \geq 0$ is a monotonically non-increasing function of $A$ by showing that $\frac{\partial}{\partial A} {\mathcal{J}}(\mathcal{P}^*) \leq 0, \forall A \geq 0$ since the latter implies the former by the mean value theorem. We will assume, without loss of generality, that $C = 1/N$, which simply means that the processing cost is the same for any voxel and the total cost when all voxels are selected is normalized to unity. To facilitate the derivations we will use the right continuous Heaviside function $H(\cdot)$ defined as $H(x)=1, x \geq 0$ and $H(x)=0, x < 0$ and the following notation $\mathcal{A}_i = A (U_i + \lambda^* /N)$.  With this notation, solution~\eqref{eqn:technicalSolution} can be written concisely as follows:
\begin{align}
p_i^* = H(\mathcal{A}_i)H(\Ph-\mathcal{A}_i)\mathcal{A}_i + \Ph(1-H(\Ph-\mathcal{A}_i))
\end{align}
Since $\mathcal{P}^*$ is the solution of~\eqref{eqn:optProblem} and it thus satisfies all the constraints we note that  ${\mathcal{J}}(\mathcal{P}^*) = -\sum_{i=1}^N p_i^* U_i$. Taking these remarks into account we can rewrite the objective function as:
\begin{align}
{\mathcal{J}}(\mathcal{P}^*) = -\sum_{i=1}^N U_i[
H(\mathcal{A}_i)&H(\Ph-\mathcal{A}_i)\mathcal{A}_i \nonumber\\
&+\Ph(1-H(\Ph-\mathcal{A}_i))]
\end{align}
and the constraint ${\mathcal{P}^*}^T \bC = C_{\ave}$:
\begin{align} \label{eqn:constraint_short}
\frac{1}{N} \sum_{i=1}^N [
H(\mathcal{A}_i)H(\Ph-\mathcal{A}_i)\mathcal{A}_i +
\Ph(1-H(\Ph-\mathcal{A}_i))] = C_{\ave}.
\end{align}
The differentiation of ${\mathcal{J}}(\mathcal{P}^*)$ results in
\begin{align} \label{eqn:J_derivative_D_A}
\frac{\partial}{\partial A}&\mathcal{J}(\mathcal{P}^*) =
-\sum_{i=1}^N U_i \frac{\partial \mathcal{A}_i}{\partial A} [
\Ph\delta(\mathcal{A}_i-\Ph) \nonumber\\
&+ \delta(\mathcal{A}_i)
H(\Ph-\mathcal{A}_i)\mathcal{A}_i - \delta(\mathcal{A}_i-\Ph)
H(\mathcal{A}_i)\mathcal{A}_i \nonumber\\
&+ H(\mathcal{A}_i)H(\Ph-\mathcal{A}_i)].
\end{align}
Consider $\Ph\delta(\mathcal{A}_i-\Ph) + \delta(\mathcal{A}_i)
H(\Ph-\mathcal{A}_i)\mathcal{A}_i - \delta(\mathcal{A}_i-\Ph)
H(\mathcal{A}_i)\mathcal{A}_i$. It is straightforward to see that this expression
is 0 whenever $\mathcal{A}_i < 0$, $0<\mathcal{A}_i < \Ph$ and
$\mathcal{A}_i > \Ph$. Consider the case $\mathcal{A}_i \rightarrow
\Ph$:
\begin{align}
\lim_{\mathcal{A}_i \rightarrow \Ph} &(\Ph\delta(\mathcal{A}_i-\Ph) +
\delta(\mathcal{A}_i) H(\Ph-\mathcal{A}_i)\mathcal{A}_i \nonumber\\
&-
\delta(\mathcal{A}_i-\Ph) H(\mathcal{A}_i)\mathcal{A}_i) \nonumber\\
&= \lim_{\mathcal{A}_i
\rightarrow \Ph} \Ph (\delta(\mathcal{A}_i-\Ph) - \delta(\mathcal{A}_i-\Ph)) =
0.
\end{align}
Consider the case $\mathcal{A}_i \rightarrow 0$:
\begin{align}
\lim_{\mathcal{A}_i \rightarrow 0} &(\Ph \delta(\mathcal{A}_i-\Ph) +
\delta(\mathcal{A}_i) H(\Ph-\mathcal{A}_i)\mathcal{A}_i \nonumber\\
&-
\delta(\mathcal{A}_i-\Ph) H(\mathcal{A}_i)\mathcal{A}_i) = \lim_{\mathcal{A}_i
\rightarrow 0} \delta(\mathcal{A}_i) \mathcal{A}_i = 0.
\end{align}
The last equality is one of the fundamental properties of the Delta
function. Thus we see that for any value of $\mathcal{A}_i$
the expression just analyzed is zero and the derivative
$\frac{\partial}{\partial A}{\mathcal{J}}(\mathcal{P}^*)$
is simply:
\begin{align}
\frac{\partial}{\partial A}{\mathcal{J}}(\mathcal{P}^*) =
-\sum_{i=1}^N U_i \frac{\partial \mathcal{A}_i}{\partial A}
H(\mathcal{A}_i)H(\Ph-\mathcal{A}_i).
\end{align}
Taking into account the fact that $\lambda^*$ is a function of $A$
and calculating the derivative
\begin{align}
\frac{\partial \mathcal{A}_i}{\partial A} =
\frac{\partial}{\partial
A} A (U_i + \lambda^* /N) = [U_i + \lambda^*/N] + \frac{\partial \lambda^*}{\partial A} \frac{A}{N},
\end{align}
we can rewrite the derivative $\frac{\partial}{\partial
A}{\mathcal{J}}(\mathcal{P}^*)$ as follows:
\begin{align} \label{eqn:J_derivative_D_lambda}
\frac{\partial}{\partial A}{\mathcal{J}}(\mathcal{P}^*) &= -
\sum_{i=1}^N H(\mathcal{A}_i)H(\Ph-\mathcal{A}_i)U_i [(U_i +
\lambda^*/N) \nonumber\\
&+ \frac{\partial \lambda^*}{\partial A} A/N].
\end{align}
To derive $\frac{\partial \lambda^*}{\partial A}$ we use the
implicit differentiation of the constraint
equation~\eqref{eqn:constraint_short} which, upon differentiating
its both sides and simplifying it using the same logic that was used to simplify~\eqref{eqn:J_derivative_D_A}, becomes:
\begin{align}
\sum_{i=1}^N \frac{H(\mathcal{A}_i)H(\Ph-\mathcal{A}_i)}{N} \left[\left(U_i +
\frac{\lambda^*}{N}\right) + \frac{\partial \lambda^*}{\partial A} \frac{A}{N}\right] = 0.
\end{align}
After grouping and moving the terms we obtain:
\begin{align}
\frac{\partial \lambda^*}{\partial A} &= -\frac{N \sum_{i=1}^N
H(\mathcal{A}_i)H(\Ph-\mathcal{A}_i)(U_i + \lambda^*/N)}{A
\sum_{i=1}^N H(\mathcal{A}_i)H(\Ph-\mathcal{A}_i)} \nonumber\\
&= -\frac{N \sum_{i=1}^N H(\mathcal{A}_i)H(\Ph-\mathcal{A}_i)U_i}{A
\sum_{i=1}^N H(\mathcal{A}_i)H(\Ph-\mathcal{A}_i)} - \lambda^*/A.
\end{align}
Inserting this into~\eqref{eqn:J_derivative_D_lambda} produces:
\begin{align}
\frac{\partial {\mathcal{J}}}{\partial A}(\mathcal{P}^*) &=
-\sum_{i=1}^N H(\mathcal{A}_i)H(\Ph-\mathcal{A}_i) \nonumber\\
&\times\left[U_i^2
-\frac{\sum_{j=1}^N H(\mathcal{A}_j)H(\Ph-\mathcal{A}_j)U_i
U_j}{\sum_{j=1}^N H(\mathcal{A}_j)H(\Ph-\mathcal{A}_j)}\right] \nonumber\\
&= -\sum_{i=1}^N \frac{ H(\mathcal{A}_i)H(\Ph-\mathcal{A}_i) }{\sum_{k=1}^N H(\mathcal{A}_k)H(\Ph-\mathcal{A}_k)} \nonumber\\
&\times\left[\sum_{j=1}^N U_i^2H(\mathcal{A}_j)H(\Ph-\mathcal{A}_j) \right. \nonumber\\
&-\left.\sum_{j=1}^N H(\mathcal{A}_j)H(\Ph-\mathcal{A}_j)U_i
U_j\right].
\end{align}
Taking into account that
\begin{align}
&\sum_{i=1}^N\sum_{j=1}^N
H(\mathcal{A}_i)H(\Ph-\mathcal{A}_i)H(\mathcal{A}_j)H(\Ph-\mathcal{A}_j)U_i^2 \nonumber\\
&= \sum_{i=1}^N\sum_{j=1}^N
H(\mathcal{A}_i)H(\Ph-\mathcal{A}_i)H(\mathcal{A}_j)H(\Ph-\mathcal{A}_j)U_j^2,
\end{align}
we can finally write:
\begin{align}
\frac{\partial}{\partial A}{\mathcal{J}}(\mathcal{P}^*) &=
-\sum_{i=1}^N\sum_{j=1}^N \frac{
H(\mathcal{A}_i)H(\Ph-\mathcal{A}_i)H(\mathcal{A}_j)H(\Ph-\mathcal{A}_j)}{2\sum_{k=1}^N
H(\mathcal{A}_k)H(\Ph-\mathcal{A}_k)} \nonumber\\
&\times\left[ U_i^2 + U_j^2 - 2U_i U_j\right] \nonumber\\
&=-\sum_{i=1}^N\sum_{j=1}^N \frac{
H(\mathcal{A}_i)H(\Ph-\mathcal{A}_i)H(\mathcal{A}_j)H(\Ph-\mathcal{A}_j)}{2\sum_{k=1}^N
H(\mathcal{A}_k)H(\Ph-\mathcal{A}_k)} \nonumber\\
&\times (U_i - U_j)^2.
\end{align}
The last expression is clearly non-positive $\forall A$.
\end{IEEEproof}

\section{Proof of proposition~\ref{prop:2}} \label{app:ProofOfProposition2}

\begin{IEEEproof}
Using the definition of $\varphi(\lambda^*)$ in the left hand side of~\eqref{eqn:constraint_short}, differentiating it with respect to $\lambda^*$ and using simplification methodology applied in Proposition~\ref{prop:1} we arrive at:
\begin{align}
\frac{\partial}{\partial \lambda^*}\varphi(\lambda^*) =
C\sum_{i=1}^N \frac{\partial \mathcal{A}_i}{\partial \lambda^*}
H(\mathcal{A}_i)H(\Ph-\mathcal{A}_i).
\end{align}
Taking into account that
\begin{align}
\frac{\partial \mathcal{A}_i}{\partial \lambda^*} =
\frac{\partial}{\partial
\lambda^*} A (U_i + \lambda^* C) = A C,
\end{align}
It is clear that $\frac{\partial}{\partial \lambda^*}\varphi(\lambda^*) \geq 0, \forall \lambda^*$ and the claim of the proposition follows.
\end{IEEEproof}

{\color{black}
\section*{Acknowledgment}

The authors would like to thank Colm Elliott, Dante De Nigris and Dr. D.L. Collins for their insightful comments and for helpful discussions.
}

\bibliographystyle{IEEEtran}
\bibliography{IPMI_pixel_sel}

\end{document}